\documentclass[10pt,journal,compsoc]{IEEEtran}
\ifCLASSOPTIONcompsoc
  \usepackage[nocompress]{cite}
\else
  \usepackage{cite}
\fi
\ifCLASSINFOpdf
  \usepackage[pdftex]{graphicx}
\else
\fi
\usepackage{amsmath}
\usepackage{array}
\usepackage{url}

\usepackage{booktabs} 

\usepackage{amssymb} 
\usepackage{stackrel} 

\usepackage[multiple]{footmisc} 

\usepackage[ruled,vlined]{algorithm2e} 

\usepackage{subfigure} 

\usepackage{thmbox} 
\newtheorem[M]{definition}{Definition} 

\newcommand{\squishlist}{
	\begin{list}{$\bullet$}
		{ \setlength{\itemsep}{0pt}      \setlength{\parsep}{3pt}
			\setlength{\topsep}{3pt}       \setlength{\partopsep}{0pt}
			\setlength{\leftmargin}{1.5em} \setlength{\labelwidth}{1em}
			\setlength{\labelsep}{0.5em} } }
	
	\newcommand{\squishlisttwo}{
		\begin{list}{$\bullet$}
			{ \setlength{\itemsep}{0pt}    \setlength{\parsep}{0pt}
				\setlength{\topsep}{0pt}     \setlength{\partopsep}{0pt}
				\setlength{\leftmargin}{2em} \setlength{\labelwidth}{1.5em}
				\setlength{\labelsep}{0.5em} } }
		
		\newcommand{\squishend}{
		\end{list}  }

\begin{document}
%
\title{Multivariate Data Explanation by Jumping Emerging Patterns Visualization}
%
%
%
%

\author{M\'ario~Popolin~Neto
        and~Fernando~V.~Paulovich,~\IEEEmembership{Member,~IEEE,}
\IEEEcompsocitemizethanks{\IEEEcompsocthanksitem M. Popolin Neto is with Federal Institute of S\~ao Paulo (IFSP) and University of S\~ao Paulo (USP), Brazil.\protect\\
E-mail: mariopopolin@ifsp.edu.br
\IEEEcompsocthanksitem F. V. Paulovich is with Eindhoven University of Technology (TU/e), the Netherlands.
\protect\\
E-mail: f.paulovich@tue.nl}}

%
%

\markboth{}%
{Popolin Neto \MakeLowercase{\textit{et al.}}: Multivariate Data Explanation by Jumping Emerging Patterns Visualization}
%



\IEEEtitleabstractindextext{%
\begin{abstract}
Multivariate or multidimensional visualization plays an essential role in exploratory data analysis by allowing users to derive insights and formulate hypotheses. Despite their popularity, it is usually users' responsibility to (visually) discover the data patterns, which can be cumbersome and time-consuming. Visual Analytics (VA) and machine learning techniques can be instrumental in mitigating this problem by automatically discovering and representing such patterns. One example is the integration of classification models with (visual) interpretability strategies, where models are used as surrogates for data patterns so that understanding a model enables understanding the phenomenon represented by the data. Although useful and inspiring, the few proposed solutions are based on visual representations of so-called black-box models, so the interpretation of the patterns captured by the models is not straightforward, requiring mechanisms to transform them into human-understandable pieces of information. This paper presents \textit{multiVariate dAta eXplanation (VAX)}, a new VA method to support identifying and visual interpreting patterns in multivariate datasets. Unlike the existing similar approaches, VAX uses the concept of Jumping Emerging Patterns, inherent interpretable logic statements representing class-variable relationships (patterns) derived from random Decision Trees. The potential of VAX is shown through use cases employing two real-world datasets covering different scenarios where intricate patterns are discovered and represented, something challenging to be done using usual exploratory approaches.
\end{abstract}

}

\maketitle

\IEEEdisplaynontitleabstractindextext

%
\IEEEpeerreviewmaketitle

\IEEEraisesectionheading{\section{Introduction}\label{sec:intro}}

\IEEEPARstart{M}{ultidimensional} or multivariate visualization plays an essential role in exploratory data analysis~\cite{Keim:2010:Mastering, Sacha:2014:Knowledge}, with the literature presenting several methods leveraging visual representations to capture and represent data patterns~\cite{Liu:2017:Visualizing}. Despite their popularity, allowing users to derive insights and formulate hypotheses, multidimensional visualization methods typically put on users' shoulders the burden of discovering (visually) the existing patterns, which can be cumbersome and time-consuming, especially for complex datasets. There is a high chance of users being unable to uncover intricate relationships in the data as the number of data variables grows~\cite{Knittel:2020:Visual}, and essential information can be lost due to human cognition limitations.

To help mitigate this issue, data mining and machine learning techniques can be instrumental, supporting users in discovering patterns in the visualizations or discovering patterns in the data that are then visualized~\cite{Dang:2014:ScagExplorer}. One interesting example of the former category is the use of classification models to capture and represent patterns, acting as descriptive instead of predictive tools~\cite{Tan:2005:Introduction}. The core idea is that if a model is transparent and understandable~\cite{Liao:2020:Questioning}, for instance, through interpretability strategies~\cite{Chan:2020:Melody}, it can be used as a proxy to discover patterns in the data enabling tasks that involve the analysis of attribute-class relationships~\cite{Gleicher:2013:Expaliners, Knittel:2020:Visual}.

Although, at first glance, any solution combining classification models and interpretability strategies, such as the recent Visual Analytics (VA) approaches for model interpretation~\cite{Ming:2019:RuleMatrix, PopolinNeto:2020:ExMatrix, Zhao:2019:iForest} could be used to employ classification models as descriptive strategies, there is an important bottleneck. Such approaches are model-centered, with generalization in mind, so inter-class separation is usually well represented while intra-class information is typically lost, limiting their application to explain data. To our knowledge, a few solutions have been proposed in the visualization literature to use classification models for descriptive purposes. Such as based on Support Vector Machines~\cite{Gleicher:2013:Expaliners} and Artificial Neural Networks~\cite{Knittel:2020:Visual}. Although inspiring approaches, both are based on visual representations of so-called black-box models~\cite{Ming:2019:RuleMatrix, Castro:2019:Surrogate, Ribeiro:2016:Why}, so the interpretation of the patterns captured by the models is not straightforward, requiring mechanisms to transform them into human-understandable pieces of information.

This paper presents \textit{multiVariate dAta eXplanation (VAX)}, a new VA approach to leverage prediction models' descriptive power for multivariate data analysis. VAX was designed to capture inter, and intra-class patterns through the so-called \textit{Jumping Emerging Patterns (JEPs)}~\cite{Bamba:2015:Minimal}. JEPs are logic statements representing class relationships (patterns) that may be extracted from random Decision Trees (DTs) inferred using the notion of Random Forests (RFs)~\cite{Breiman:2001:RandomForest, Biau:2016:Arandom}. JEPs are high (class) discriminative, diversified, and expressive patterns~\cite{Loyola:2020:AReview, Vico:2018:Anoverview} depicting multi-class multi-attribute relationships in an inherent interpretable format. VAX aggregates JEPs to reduce their number and displays them using a compact matrix metaphor and Dimensionality Reduction (DR) layouts, supporting different analytical tasks involving pattern and data content (clusters and outliers) analysis, revealing intricate and complex information that may be challenging to discover using usual exploratory approaches.

In summary, the main contributions of this paper are:

\squishlist
    \item A new strategy for JEPs selection and aggregation from random DTs that helps to summarize large sets of patterns while representing the entire data set;
    
    \item A new method for JEPs visualization, where a matrix metaphor is used to display patterns as rows, variables as columns, and data information through histograms in the cells; and
    
    \item A DR similarity map for analyzing data instances from the perspective of the discovered patterns, composing an analytical cycle that goes from data to patterns and from patterns to data.
\squishend

The remainder of the paper is organized as follows. Sec.~\ref{sec:related} covers the literature in classification model visualization for descriptive analysis, discussing the current limitations and positioning our solution. Sec.~\ref{sec:method} details our proposed approach, showing how JEPs are extracted, aggregated, and visualized. Sec.~\ref{sec:results} presents two different use cases explaining how to use our solution for data explanation. Finally, Sec.~\ref{sec:limitations} lists our limitations and Sec.~\ref{sec:conclusions} outlines our conclusions and future work.

\section{Related Work}
\label{sec:related}

There is extensive literature on the visual analysis of multivariate data (see~\cite{Liu:2017:Visualizing} for a survey), granting a myriad of methods and systems, including Dimension Reduction (DR) strategies~\cite{Laurens:2008:tSNE, Jeong:2009:iPCA, Paulovich:2010:Two, Turkay:2012:Representative, McInnes:2018:UMAP, Fujiwara:2020:ccPCA, Nonato:2019:Multidimensional}, clustering approaches~\cite{Cao:2011:Dicon, Yuan:2013:Dimension, Bernard:2014:Visual}, statistical correlation methods~\cite{Piringer:2008:Quantifying, Malik:2012:Correlative, Zhang:2015:Visual, Goodwin:2016:Visualizing, May:2011:Guiding}, to mention a few. Typically, these methods and systems focus on representing data to allow users to visually capture and interpret data patterns. In contrast, VAX uses classification models to capture data patterns that are then displayed through a visual representation, leveraging the capability of classifiers to help users discover and understand such patterns.

Visual representations of classification models have been typically used to support building~\cite{Ming:2020:Proto, Elzen:2011:BaobabView, Teoh:2003:PaintingClass}, comparing~\cite{Wang:2022:Learning, Talbot:2009:EnsembleMatrix}, and explaining/interpreting prediction models~\cite{PopolinNeto:2020:ExMatrix, Zhao:2019:iForest, Ming:2019:RuleMatrix, Ribeiro:2018:Anchors, Ribeiro:2016:Why, Lundberg:2017:SHAP, Mazumdar:2021:Random}. However, in the visualization literature, the idea of using classification models specifically as descriptive tools instead of predictive engines, using them as proxies to understand or describe multivariate data, is relatively new. Here, we divide the existing approaches into two groups: (1) model specific~\cite{Gleicher:2013:Expaliners, Knittel:2020:Visual, Li:2022:Incorporation, Liu:2022:RankAxis}; and (2) logic rules and emerging patterns~\cite{Dong:1999:EP, Novak:2009:Supervised, Vico:2018:Anoverview, Loyola:2020:AReview}.

\subsection{Model Specific}

Classification models combined with visual representations have been used for descriptive purposes. Explainers~\cite{Gleicher:2013:Expaliners} is one example. It employs DR layouts created using linear functions from Support Vector Machines (SVM) models. Explainers allows data analysis using such linear functions to interpret DR layout patterns. It is an inspiring and pioneer approach but is limited to present patterns resulting from linear combinations of up to three variables, missing patterns in more complex non-linear associations. RankAxis~\cite{Liu:2022:RankAxis} also employs linear SVM models, ranking data instances and obtaining attribute weights that serve as input to build DR layouts. Despite being a promising technique, RankAxis may present disparities in ranking and DR results and is also limited in providing linear approximations for non-linear data relationships. In our approach, the patterns extracted can involve more than three variables and represent non-linear relationships among instances and their classes. To allow that, we use the concept of \textit{Jumping Emerging Patterns (JEPs)}~\cite{Dong:1999:EP, Novak:2009:Supervised, Loyola:2020:AReview} to extract patterns with different variable combinations. Furthermore, patterns are also used to create DR layouts, enabling analysis involving patterns and data instances.

Another example is Visual Neural Decomposition (VND)~\cite{Knittel:2020:Visual}. VND enables multivariate data explanation by visually presenting Artificial Neural Network hidden node weights through stacked bars to depict the relations between variable ranges and classes (e.g., class A with a threshold probability). The nodes are organized in cards containing variables ordered by importance to a node. Although VND can show non-linear relationships among instances and a particular class, the captured patterns' complexity is bounded by the simple neural network architecture (one hidden layer) employed to allow interpretability. Also, their class-specific visualizations can make the analysis of multiple classes difficult. A single hidden layer network has also been used as a customized classification loss function for creating DR layouts~\cite{Li:2022:Incorporation}. Such enhanced layouts can reveal clusters and outliers, where data instances selections are explained using SHAP~\cite{Lundberg:2017:SHAP}, so lacking the ability to convey information involving multiple variables combinations. In our approach, since we use inherent interpretable logic statements from JEPs and a matrix visual representation, complex multiple class relations can be discovered and concisely displayed. Other positive aspects of our approach are capturing patterns with maximum confidence (and statistical significance), so that a specific class pattern does not support instances from another class (not limited to a class probability threshold), and extracting patterns that involve multiple variable combinations.

The above-mentioned approaches are inspiring but lack the explanatory power of JEPs~\cite{Dong:1999:EP, Novak:2009:Supervised, Vico:2018:Anoverview, Loyola:2020:AReview}, where one of the main goals is to obtain patterns for data explanation.

\subsection{Logic Rules and Emerging Patterns}

JEPs consist of class-associated relational statements among variables, providing class differentiation and emerging trends~\cite{Dong:1999:EP, Novak:2009:Supervised, Vico:2018:Anoverview, Anwar:2017:Survey}. VAX extracts JEPs using random DTs inducted based on the concept of Random Forests (RFs)~\cite{Borroto:2015:Finding, Loyola:2019:Fusing, Loyola:2020:AReview}, obtaining many diversified and expressive patterns~\cite{Vico:2018:Anoverview, Loyola:2020:AReview}. These patterns are then post-processed by a selection and aggregation strategy to reduce their number before visualization. JEPs are a particular case of a more general concept called \textit{Emerging Patterns (EPs)}~\cite{Vico:2018:Anoverview} where the confidence is maximum; the mined patterns are class-exclusive, not supporting instances of different classes.

Visualization is a powerful component when analyzing a phenomenon through data patterns~\cite{Novak:2009:Supervised, Loyola:2020:AReview}. In this context, two crucial aspects should be supported, the ability to check patterns content (e.g., data distribution) and to support multi-class investigation (more than two classes)~\cite{Novak:2009:Supervised}. Some approaches address the visualization of pattern properties (e.g., support) through visual markers~\cite{Novak:2009:Supervised} but lack patterns' content representation. Visualizing Subgroup Distribution (VSD)~\cite{Gamberger:2002:Subgroupvisualization} presents patterns as line plots for continuous variables and binary class problems (two classes). Despite being intuitive, VSD~\cite{Gamberger:2002:Subgroupvisualization} is not suitable for multi-class domains, and multiple variables visualization is an issue~\cite{Novak:2009:Supervised}. Our approach satisfies both content and multi-class requirements, presenting patterns' content using histograms and classes mapped to categorical colors.

JEPs can be seen as descriptive logic rules~\cite{Vico:2018:Anoverview, Novak:2009:Supervised}, and disjoint and consistent logic rules are deemed interpretable~\cite{Miranda:2021:Preventing, Lakkaraju:2016:Interpretable, Frunkranz:2012:Foundations}, therefore popular in VA solutions for model interpretability~\cite{PopolinNeto:2020:ExMatrix, Ming:2019:RuleMatrix, Ribeiro:2018:Anchors, Guidotti:2018:Local, Lakkaraju:2016:Interpretable}. Such solutions usually support global and local model interpretation to explain the model itself and its decisions~\cite{Du:2018:Techniques}. ExMatrix~\cite{PopolinNeto:2020:ExMatrix}  and RuleMatrix~\cite{Ming:2019:RuleMatrix} are two examples of such techniques. ExMatrix focuses on supporting the interpretation of RF models, displaying rules extracted from the random Decision Trees (DT) in a matrix visualization. Although ExMatrix has shown to be a good solution for analyzing RF models, no information is conveyed about the data, so limited if the goal is to understand data patterns. RuleMatrix, a matrix visualization to depict logic rules from surrogate models, presents some information about the data. However, it does not support the analysis of data instances in micro and macro contexts. It presents information through global histograms but does not allow the interpretation of the data represented by the logic rules or individual data instances. Despite great solutions, ExMatrix and RuleMatrix are model-centered, where rules are primarily used in explaining models, lacking essential capabilities if the goal is to interpret multivariate data. Our solution also leverages a matrix metaphor, but our goal is to support explanations of multivariate data, not models. It focuses on descriptive rules/patterns, supporting the explanation of groups (e.g., clusters) or distinct data instances (e.g., outliers), offering different visual representations to navigate between patterns and instances.

Finally, another VA approach that could be potentially considered for description tasks is iForest~\cite{Zhao:2019:iForest}. Similar to ExMatrix, iForest was developed for the interpretation of RF models. However, unlike ExMatrix, it only supports local interpretability, that is, the explanation of individual instances predictions. For descriptive tasks, in which the goal is the interpretation of (complete) datasets, global interpretability needs to be supported so that it is possible to understand data through the average behavior of a model.

\section{Method}
\label{sec:method}

This section presents \textit{multiVariate dAta eXplanation (VAX)}, a new multidimensional data explanation approach that combines JEPs visualization and DR layouts to support data pattern discovery and interpretation. To reach our general objective, the proposed visual representations implement five automated data insights presented by Law et al.~\cite{Law:2020:Characterizing} (Table~\ref{tab:GoalsInsightsTypes}).

\begin{table}[htb]
    \caption{Automated data insights implemented to reach our objective.}
    \label{tab:GoalsInsightsTypes}
    \scriptsize
    \centering
    \begin{tabular}{ c p{7.25cm} } 
        
        \toprule
        \multicolumn{2}{c}{Insight Type~\cite{Law:2020:Characterizing}} \\
        
        \midrule        
        
        \textbf{I1} & \textbf{Visual motifs.} Unique/special/specific patterns, being but not only custom visual metaphors, representing a particular notion/structure on data. \\ 
        \textbf{I2} & \textbf{Distribution.} Variables values distribution, such as histogram plots. \\
        \textbf{I3} & \textbf{Cluster.} Instances group, like a set of points relative closed to each other on a scatter plot. \\
        \textbf{I4} & \textbf{Outlier.} Particular instance with distinct variables values to the distribution, such as an instance relative apart from other instances in a scatter plot.   \\
        \textbf{I5} & \textbf{Compound fact.} Meaningful composition of two or more insights types. \\
        
        \bottomrule 
        
    \end{tabular}
\end{table}

By implementing these insight types, VAX supports the needed elements for exploratory tasks inside the visual analytics process~\cite{Keim:2010:Mastering, Sacha:2014:Knowledge}. We implement \textbf{I1 - Visual motifs} and \textbf{I2 - Distribution} using a matrix-like visual metaphor adopting matrix visualization guidelines~\cite{Chen:2004:MatrixVisualization, Wu:2008:MatrixVisualization}. Since descriptive patterns (JEPs) are our objects of analysis, we represent them as rows (\textbf{I1}), data variables as columns, and fill matrix cells with global and local histograms (\textbf{I2}). This arrangement combines the strengths of model explanation approaches~\cite{PopolinNeto:2020:ExMatrix, Ming:2019:RuleMatrix}, improving them for data interpretation instead of model analysis. The insight types \textbf{I3 - Cluster} and \textbf{I4 - Outlier} are implemented using a DR technique, creating a similarity map as viewed through the JEPs lens, where clusters (\textbf{I3}) and outliers (\textbf{I4}) can be observed. Finally, \textbf{I5 - Compound fact} is reached through combining the different proposed visual representations where clusters and outliers (\textbf{I3} and \textbf{I4}) can be explained by visualizing the patterns along with variables distributions (\textbf{I1} and \textbf{I2}).

\begin{figure*}[h]
    \centering
    \includegraphics[width=\linewidth]{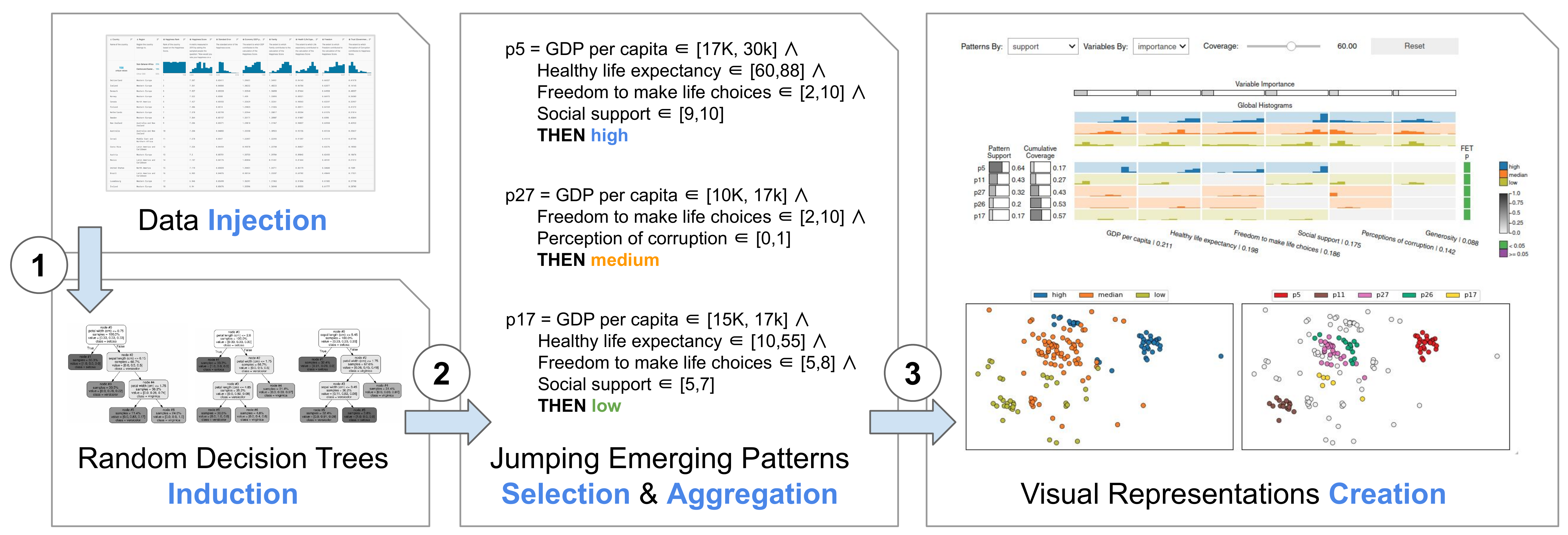}
    \caption{VAX data explanation pipeline. \textbf{1}: Random Decision Trees are inducted based on the entire dataset. \textbf{2}: Jumping Emerging Patterns (JEPs) are extracted from the trees and then selected and aggregated following a well-defined strategy. \textbf{3}: JEPs are visualized using a matrix-like visual metaphor composed of global and local histograms (insights \textbf{I1} and \textbf{I2}), and similarity maps are built to show the instances' similarity relationships from the patterns' perspective  (insights \textbf{I3} and \textbf{I4}). From the matrix visualization, JEPs can be further analyzed by inspecting the supported instances on the map  (insights \textbf{I5}). From the map, instances can be further investigated by inspecting the JEPs which support them (\textbf{I5}).}
    \label{fig:Pipeline}
\end{figure*}

The proposed VAX pipeline is presented in Fig.~\ref{fig:Pipeline}. Initially, random Decision Trees (DTs) are inducted considering the entire dataset (\textbf{1}). JEPs are extracted from the DTs, and the more relevant patterns are selected and aggregated following a well-defined strategy (\textbf{2}). The resulting aggregated patterns are then visualized using a matrix metaphor, and the similarity relationships among data instances are represented using a DR layout with the similarity among instances defined by the patterns (\textbf{3}). This allows users to explore how patterns are related to instances and how instances are connected with patterns. Next, we detail how the JEPs matrix and the similarity map are created.

\subsection{Jumping Emerging Patterns Matrix}

Before explaining the construction of the JEPs matrix, we first introduce and formalize the concept of Emerging Patterns (EPs)~\cite{Dong:1999:EP, Vico:2018:Anoverview, Loyola:2020:AReview} and Jumping Emerging Patterns (JEPs)~\cite{Vico:2018:Anoverview, Bamba:2015:Minimal} and how they are extracted from datasets.

\subsubsection{Definitions}

An EP is a descriptive rule combining dataset variables using relational operators to express emerging relationships (i.e., trends and differentiating behavior) in data~\cite{Novak:2009:Supervised, Vico:2018:Anoverview}. In formal terms, given a dataset ${X=\{x_1, \ldots, x_N\}}$ with each instance represented by ${V =\{v_1, \ldots, v_M\}}$ variables. A pattern $p$ is a conjunction of selectors (logical complex), each one defining a relational statement in the form of $v_i \; \otimes \; S_i$, where $S_i$ consists of one or more possible values (or a range) for the variable $v_i \in V$ and the relational operator $\otimes$ is $=, \neq,\in,\notin,>,<,\geq,$ or $\leq$~\cite{Michalski:1982:Revealing, Vico:2018:Anoverview}, that is

\begin{equation}
    p = [v_1 \; \otimes \; S_1] \; \land \;  [v_2 \; \otimes \; S_2] \; \land \; \ldots \; \land \;[v_{M} \; \otimes \; S_{M}]
\end{equation}

An instance $x_i \in X$ (a.k.a transaction, sample, example, or item) is said to be supported by a pattern $p$ if it satisfies all selectors of $p$~\cite{Michalski:1982:Revealing, Dong:1999:EP, Vico:2018:Anoverview}. Moreover, given $X_a \subset X$ and $X_b \subset X$, with $X_a \cap X_b = \varnothing$, two disjoint partitions (i.e., sub-datasets) of $X$, a pattern $p$ is considered an EP in $X_b$ given $X_a$ if the percentage of instances supported by $p$ in $X_b$ divided by the percentage of instances supported in $X_a$ is larger than a threshold ($\geq 1$), that is, if the pattern $p$ is more ``valid'' for the instances in $X_b$ than in $X_a$. This ratio is called Growth Rate~\cite{Michalski:1982:Revealing, Dong:1999:EP, Vico:2018:Anoverview}, and it is defined by

\begin{equation}
  GR(X_a, X_b, p) = \begin{cases}
    0 & \begin{split}
            \text{If } & Supp(X_a, p) = 0~\land \\
                      & Supp(X_b, p) = 0
        \end{split}\\
    \infty & \begin{split}
                \text{If } & Supp(X_a, p) = 0~\land \\
                          & Supp(X_b, p) \neq 0
            \end{split}\\
    \frac{Supp(X_b,p)}{Supp(X_a, p)} & \text{Otherwise}
  \end{cases}
  \label{eq:GR}
\end{equation}

where $Supp(X_i, p)$ is the support of $p$ in the dataset $X_i \subset X$, given by

\begin{equation}
  Supp(X_i, p) = \frac{count(X_i, p) }{ |X_i| }
  \label{eq:SUP}
\end{equation}

with $count(X_i, p)$ the number of instances from $X_i$ supported by $p$, and $|X_i|$ the cardinality of $X_i$~\cite{Dong:1999:EP, Novak:2009:Supervised, Vico:2018:Anoverview}. The EP core idea is to discover patterns whose support increases (Growth Rate) from a partition $X_a$ to a partition $X_b$~\cite{Dong:1999:EP, Novak:2009:Supervised, Vico:2018:Anoverview}. For class-labeled datasets, where each instance $x_i \in X$ is associated with a class $y_i \in C = \{ c_{1},...,c_{J~\geq~2} \}$, the partitions can be composed by instances associated with the same label~\cite{Vico:2018:Anoverview}. In this way, for binary-class datasets ($J = 2$), $X_a$ will contain instances of one class, while $X_b$ will contain the other class instances. For multi-class datasets ($J > 2$), an \textit{One-vs-All} strategy can be used~\cite{Vico:2018:Anoverview}, and $X_b$ will contain the instances of a particular class, and $X_a$ will contain the instances of all remaining classes. In the remainder of this paper, the terms partition or class have the same meaning and will be used interchangeably.

\begin{figure}[h]
    \centering
    \includegraphics[width=\columnwidth]{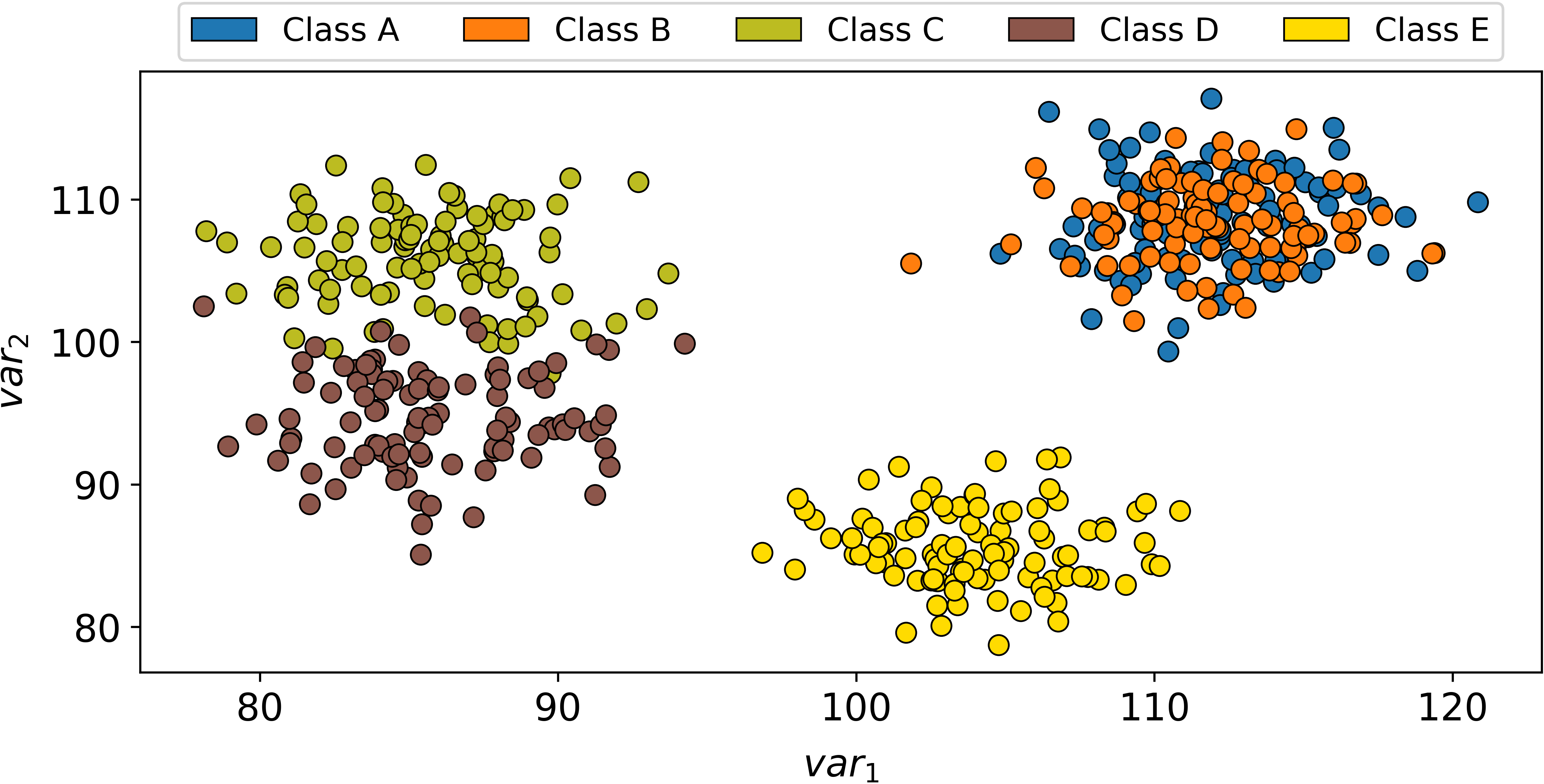}
    \caption{The synthetic dataset $X_{S}$ used to illustrate our approach. It covers different scenarios where EPs can be used to describe a dataset, including instances belonging to strongly mixed  (A and B),  adjacent (C and D), and well-separated classes (E).}
    \label{fig:SD-SP}
\end{figure}

For illustration, we use the synthetic dataset $X_{S}$ presented in Fig.~\ref{fig:SD-SP}, covering different scenarios. $X_{S}$ is composed of $500$ instances, described by $2$ real variables ($var_1, var_2$), and equally distributed among $5$ classes ($J = 5$), $A$, $B$, $C$, $D$, and $E$ ($100$ instances per class). Three clusters can be spotted in $X_{S}$, where circles denote instances and color reflects their class. Instances of class $A$ (blue) and $B$ (orange) strongly overlap, instances of class $C$ (green) and $D$ (brown) are adjacent, and instances of class $E$ (yellow) are completely isolated.

Considering $X_{A} \subset X_{S}$ the partition containing all instances of class $A$ (the same applying for the other classes $B$, $C$, $D$, and $E$), the pattern ${p_{ex} = [var_1 \in [96,120]] \; \land \; [var_2 \in [78,95]]}$ is a EP for class $E$ instances (partition $X_{E}$) given the other classes. The pattern $p_{ex}$ Growth Rate equals to $GR(\{X_A \cup X_B \cup X_C \cup X_D\}, X_E, p_{ex}) = \infty$, since $Supp(X_{E}, p_{ex}) = 1.0$ and $Supp(\{X_A \cup X_B \cup X_C \cup X_D\}, p_{ex}) = 0$. The pattern $p_{ex}$ supports all class $E$ instances ($count(X_E, p_{ex}) = |X_{E}| = 100$), but none instances from classes $A$, $B$, $C$, or $D$ (partitions $X_A, X_B, X_C$, or $X_D$) as $count(\{X_A \cup X_B \cup X_C \cup X_D\}, p_{ex}) = 0$ and $|\{X_A \cup X_B \cup X_C \cup X_D\}| = 400$.

There are different types of EPs based on the relationships between variables~\cite{Vico:2018:Anoverview, Loyola:2020:AReview}. In this paper, we are interested in the so-called \textit{Jumping Emerging Patterns (JEPs)} given their discriminative power among classes~\cite{Vico:2018:Anoverview, Bamba:2015:Minimal}. A JEP is an EP observed in only one class given all other classes, defined by

\vspace{0.35cm}
\begin{definition}[Jumping Emerging Pattern -- JEP]
Let $X_a \subset X$ and $X_b \subset X$ be two disjoint partitions of $X$, that is, $X_a \cap X_b = \varnothing$, and $p \in P$ be an \textit{Emerging Pattern (EP)} of $X_b$ given $X_a$, a \textit{Jumping Emerging Pattern (JEP)} is an EP with Growth Rate equals to $\infty$, that is, ${\{p \in P | GR(X_a, X_b, p) = \infty\}}$.
\end{definition}
\vspace{0.35cm}

It is worth mentioning that JEPs also present a maximum value ($1.0$) of confidence (pattern precision)~\cite{Vico:2018:Anoverview}, defined by

\begin{equation}
  Conf(X_a, X_b, p) = \frac{ count(X_b, p) }{ count(X_a, p) + count(X_b, p) }
  \label{eq:CONF}
\end{equation}

Pattern ${p_{ex} = [var_1 \in [96,120]] \; \land \; [var_2 \in [78,95]]}$ is a JEP, since $GR(\{X_A \cup X_B \cup X_C \cup X_D\}, X_E, p_{ex}) = \infty$. Moreover, it presents confidence equals to $Conf(\{X_A \cup X_B \cup X_C \cup X_D\}, X_E, p_{ex}) = 1.0$, since $count(X_{E}, p_{ex}) = 100$ and $count(\{X_A \cup X_B \cup X_C \cup X_D\}, p_{ex}) = 0$. From this point forward, we use the synthetic dataset $X_{S}$ to help explain our approach, including JEPs extraction, selection, aggregation, and visualization.

\subsubsection{Extraction}
\label{subsec:ext}

Mining EPs is an NP-Hard problem resulting from the exponential number of candidate patterns that are generated as the number of variables grows~\cite{Wang:2004:Complexity, Li:2004:Incremental, Vico:2018:Anoverview, Loyola:2020:AReview}. In this context, EPs can be mined using different heuristics~\cite{Novak:2009:Supervised, Vico:2018:Anoverview, Loyola:2020:AReview}. One popular approach is to use Decision Tree (DT) models~\cite{Breiman:1984:Classification, Tan:2005:Introduction}. The idea is to build varied DTs using diverse (random) factors and then extract a pattern from each decision path (the path from the root to a leaf node)~\cite{Novak:2009:Supervised, Dong:1999:EP, Vico:2018:Anoverview, Loyola:2020:AReview}. One of the most powerful techniques to mine EPs, and the method we use in this paper, is the  Random Forest miner (RFm)~\cite{Borroto:2015:Finding, Loyola:2019:Fusing, Loyola:2020:AReview}. RFm was developed based on the concept of Random Forests (RFs)~\cite{Breiman:2001:RandomForest, Biau:2016:Arandom}, and it has been proved to be a reasonable strategy for obtaining diversified patterns~\cite{Borroto:2015:Finding}.

Shortly, let $V$ be the variables of a class-labeled dataset $X$. The RFm method builds $k$ unpruned DTs from $X$ selecting a random subset of variables at each internal node creation, where the variables subset size is $\log_{2}|V|$. Unlike the RF method proposed in~\cite{Breiman:2001:RandomForest}, the bagging process (random selection of training instances)~\cite{James:2013:An} is not used in RFm, thus avoiding hidden dependencies among patterns and data instances~\cite{Borroto:2015:Finding, Loyola:2020:AReview}.

Algorithm~\ref{alg:EXT} presents the extraction process, where a class-labeled dataset $X$ and the number of trees $k$ are input parameters. Function \textsc{DecisionTree()} creates a DT model~\cite{Breiman:1984:Classification, Breiman:2001:RandomForest, Tan:2005:Introduction, James:2013:An} while \textsc{ExtractPatterns()} extracts a pattern $p$ for each decision path (root to leaf node) of a given DT~\cite{Novak:2009:Supervised, Vico:2018:Anoverview, Loyola:2020:AReview}. Since our approach focuses on descriptive analysis, we set $k$ as the minimum number of trees containing enough patterns to represent all data instances (this is later discussed in Sec.~\ref{subsec:usecasei}). Also, the entire dataset $X$ is employed to build the random DT models. Unlike predictive analysis, which focuses on creating generic models and usually splits a dataset into training and testing subsets~\cite{James:2013:An}, descriptive tasks are specific to a given dataset. They do not intend to be generic outside the scope of the data under analysis but rather offer explanations of the phenomena observed in a single dataset~\cite{Gleicher:2013:Expaliners, Knittel:2020:Visual}.

\begin{algorithm}
    \SetAlgoLined
    
    \KwIn{$X$ - Class-labeled Dataset \\
          \hspace{1.1cm}$k$ - Number of Trees}
    \KwOut{$P$ - Emerging Patterns}
    
    $P \; \leftarrow \; \varnothing$\;
    
    \For{1 to k} {
    
        $DT \; \leftarrow \; DecisionTree( Dataset = X, \; Subset \; Size = \log_2 )$\;
        
        $P \; \leftarrow \; P \; \cup \; ExtractPatterns(DT)$\;
        
    }
\caption{EPs extraction using random DTs.}
\label{alg:EXT}
\end{algorithm}

One relevant factor is that the \textit{Fisher Exact Test (FET)} can be applied to compute statistical significance per pattern~\cite{Boulesteix:2003:ACART, Novak:2009:Supervised, Loekito:2009:Using}. Values above the significance p-value (usually $0.05$) imply the null-hypothesis acceptance that there is no association between the pattern and a class~\cite{Loekito:2009:Using}.

Despite the ability to provide diversified patterns~\cite{Borroto:2015:Finding, Loyola:2020:AReview}, representing assorted class and variable-range combinations, random DT models usually extract redundant patterns~\cite{Borroto:2015:Finding, Loyola:2019:Fusing}. To address this problem, we present a novel pattern aggregation strategy to avoid overloading users with repetitive information. It is next described.

\subsubsection{Selection and Aggregation}

Once all patterns are extracted from the $k$ random DT models (Algorithm~\ref{alg:EXT}), originating a set of patterns $P$, these must be selected and aggregated. Inspired by the selection method presented in~\cite{Loyola:2019:Fusing}, we analyze patterns in $P$ from the highest to the lowest support, choosing a subset of patterns with maximum confidence that does not support data instances supported by other patterns. This way, each instance is represented by only one JEP ($GR = \infty$). The main difference between our strategy and the one presented in~\cite{Loyola:2019:Fusing} is that we do not discard supplementary patterns (different patterns that support the same instances). The supplementary patterns are aggregated, not losing their information.

The selection and aggregation process is summarized in Algorithm~\ref{alg:SELAGG}. Using an iterative greedy process applied to the set of patterns $P$ (resulted from Algorithm~\ref{alg:EXT}) ordered by decreasing support, we select as a candidate the first pattern (highest support) $p_{candidate}$. If the confidence of $p_{candidate}$ equals $1.0$, it is selected as $p_{pivot}$ and aggregated with the subset of patterns $P' \subset P$ that support the same instances.

\begin{algorithm}
    \SetAlgoLined
    
    \KwIn{$P$ - Emerging Patterns}
    \KwOut{$P$ - Jumping Emerging Patterns}
    
    $SI \; \leftarrow \; \varnothing$\;
    $P \; \leftarrow \; OrderByDecreasingSupport(P)$\;
    
    \While{$p_{candidate} \; \leftarrow \; Next(P)$}{
    
        \eIf{$Confidence( p_{candidate} ) = 1.0 \;\; and $ $SupportedInstances( p_{candidate} ) \notin SI$}{
        
            $p_{pivot} \; \leftarrow \; p_{candidate}$\;
        
            $SI \; \leftarrow \; SI \; \cup \; SupportedInstances( p_{pivot} )$\;
                         
            \ForEach{ $p_{a} \in P$ } {
            
                \If{ $SupportedInstances( p_{pivot} )$ = $SupportedInstances( p_{a} )$ }{
                    
                    $p_{pivot} \; \leftarrow \; Aggregate(p_{pivot}, \; p_{a})$\;
                    
                    $Remove(p_{a}, \; P)$\;
                    
                }                
            }        
        }{
        
            $Remove(p_{candidate}, \; P)$\;            
        }   
    }     
\caption{JEPs selection and aggregation.}
\label{alg:SELAGG}
\end{algorithm}

For the aggregation procedure, without loss of generality, we assume that the patterns in $P$ have selectors in the form $v \in S$, with $S = [a, b]$ and $a$ and $b$ $\in \mathbb{R}$. In this procedure, given a pattern $p_{a} \in P'$ supporting the same instances of $p_{pivot}$, all patterns selectors for the same variable are aggregated as the intersection of the two defined real sets ($S^{p_{pivot}}$ and $S^{p_{a}}$). So, supposing $p_{pivot}$ and $p_{a}$ having selectors for a variable $v_i \in V$, the aggregation is $S_{i}^{p_{pivot}} \cap S_{i}^{p_{a}}$. For the case where the selector for a variable $v_i \in V$ is found in only one pattern ($p_{pivot}$ or $p_{a}$), the missing real set is $[min(X^{v_i}), max(X^{v_i})]$, being $min(X^{v_i})$ and $max(X^{v_i})$ the minimum and maximum values for the variable $v_i$ in $X$. After the aggregation of the patterns in $P'$ with $p_{pivot}$, the patterns in $P'$ are removed from $P$. In this way, complementary patterns turn into a single pattern.

The next candidate pattern $p_{candidate} \in P$ is then analyzed. If $p_{candidate}$ confidence value differs from $1.0$, $p_{candidate}$ is discarded, since it is not a JEP. However, if the confidence equals $1.0$, the $p_{candidate}$ supported instances are investigated. If $p_{candidate}$ supports at least one instance already supported by the previously selected and aggregated patterns, $p_{candidate}$ is discarded, removing a redundant pattern. If $p_{candidate}$ supports only instances not supported by the previously selected and aggregated patterns, it is selected as $p_{pivot}$ and aggregated with the set of patterns $P'$ that support the same instances, removing from $P$ this set of patterns after aggregation. This process is repeated until the end of $P$ is reached, selecting and aggregating complementary patterns and discarding the redundant ones.

\begin{figure}[h]
    \centering
    \includegraphics[width=\columnwidth]{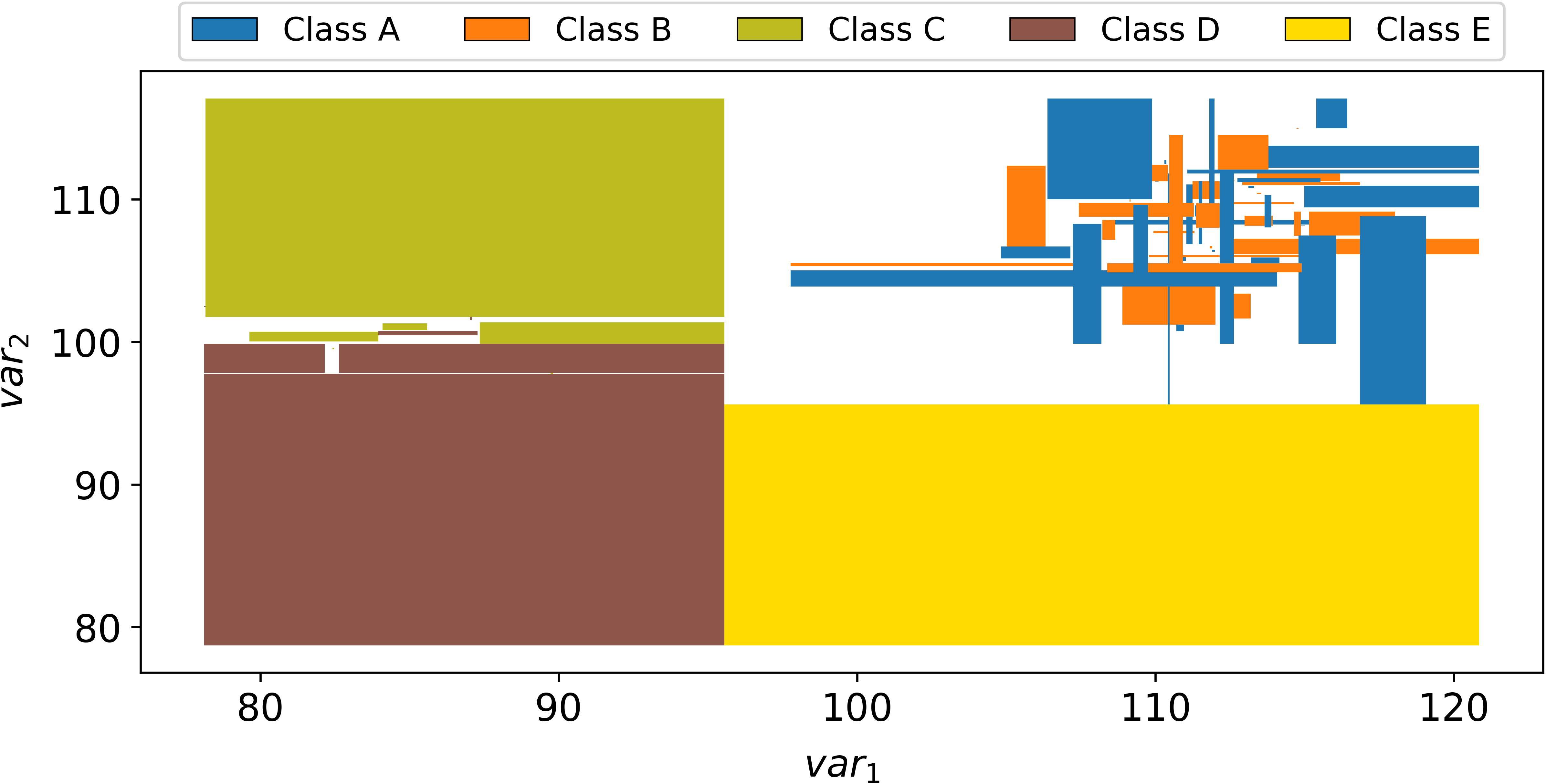}
    \caption{The regions delimited by the $67$ JEPs resulted from the selection and aggregation process using Algorithm~\ref{alg:SELAGG}. Algorithm~\ref{alg:SELAGG} takes as input $13,975$ patterns extracted using Algorithm~\ref{alg:EXT} from the synthetic dataset $X_{S}$. The number of JEPs necessary to explain a class is proportional to how complex and separated it is from the other classes.}
    \label{fig:SD-RECTS}
\end{figure}

To illustrate the selection and aggregation process, consider the execution of Algorithm~\ref{alg:EXT} to create $P$ taking the synthetic dataset $X_S$ and the number of trees $k = 128$ as inputs, resulting in $13,975$ patterns. After Algorithm~\ref{alg:SELAGG} is applied for selection and aggregation, $67$ aggregated JEPs are produced, supporting all instances from the synthetic dataset $X_{S}$. Fig.~\ref{fig:SD-RECTS} presents how these $67$ patterns segment the bi-dimensional space where $X_S$ is contained. For the instances of class $E$, the isolated groups of points, only one JEP is used to represent the entire class. For the adjacent classes $C$ and $D$, more JEPs are necessary since the overlapped border needs more patterns to describe it. For the two completed overlapped classes $A$ and $B$, several JEPs are used given the complexity to explain the differences between these classes.

In terms of computational complexity, the entire process of creating the aggregated JEPs can be split into two major components, extraction (Algorithm~\ref{alg:EXT}) and selection and aggregation (Algorithm~\ref{alg:SELAGG}). Considering the simple algorithm for DT induction, which orders the dataset variables (dimensions) before splitting the tree nodes, the complexity of inducting one tree is O($M * N \log N$), where $M$ is the number of variables and $N$ instances. If $k$ trees are created, the complexity of extraction is O($k * M * N \log N$). Given the employed greedy approach, the selection and aggregation complexity is $O(|P|^2)$, where $|P|$ is the number of extracted patterns. Since, in the worst-case scenario, the upper limit is to have $N$ leaves in a DT, the maximum number of extracted patterns is $k * N$, and the complexity of the selection and aggregation process is O($k^2 * N^2$).

This selection and aggregation process results in meaningful aggregated JEPs supporting all instances of a dataset, explaining it through high support patterns, discarding the redundant ones, and using low support patterns to explain outliers.

\subsubsection{JEPs Matrix Visualization}

\begin{figure}[h]
    \centering
    \includegraphics[width=.9\columnwidth]{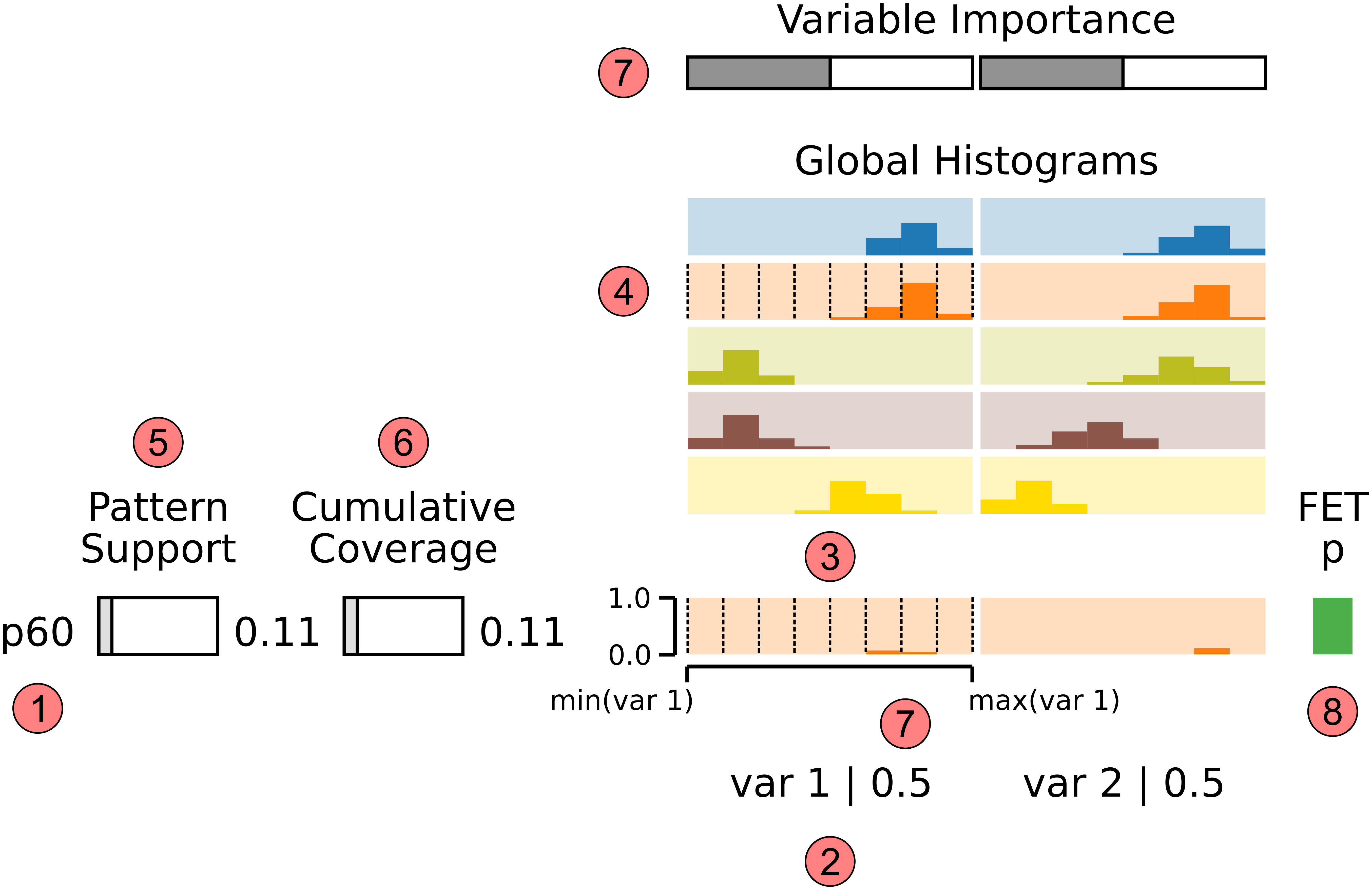}
    \caption{The matrix-like visual metaphor. \textbf{1}: JEPs are displayed as rows. \textbf{2}: Variables are arranged as columns. \textbf{3}: Cells show local normalized histograms. \textbf{4}: Global histograms (one row per class) are placed on the top, also normalized. \textbf{5}: Pattern support. \textbf{6}: Cumulative coverage assuming the matrix order (top to bottom). \textbf{7}: Variable importance. Both pattern support, cumulative coverage, and variable importance are mapped to size and brightness (grayscale). \textbf{8}: FET (Fisher Exact Test) significance value colored as green (statistically significant) or purple (not significant).}
    \label{fig:SD-MMV}
\end{figure}

Once the JEPs are selected and aggregated, we use a matrix visual metaphor for presentation (insight \textbf{I1}). Instead of a matrix, we have also considered a graph metaphor, as used to visualize association rules~\cite{10.1145/1363686.1363967, 4457213}. However, graphs do not scale well and cannot support comparisons of patterns considering the variables. Moreover, user tests have shown that rules organized into tables are easier to understand if compared to node-link arrangements~\cite{Huysmans:2011:Anempirical}. Fig.~\ref{fig:SD-MMV} presents an example of our matrix representation. It places patterns as rows (\textbf{1}) and variables as columns (\textbf{2}). Classes are mapped to categorical colors. Matrix cells (\textbf{3}) present (local) variables' histograms~\cite{Munzner:2014:Visualization} (insight \textbf{I2}) showing the variables' distributions of the instances supported by a particular pattern. Similarly, global histograms (insight \textbf{I2}) for each class are laid out on the top of the matrix (\textbf{4}). The number of bins is determined based on the Freedman-Diaconis rule~\cite{Freedman:1981:Bin, Correll:2019:Looks}, although it can also be freely defined. We have also considered using violin plots instead of histograms, but they occupy more visual space, decreasing visual scalability. If a pattern (row) does not have a selector for a variable (column), the respective cell is left empty (no histogram) by default. A local histogram can also be displayed if the user wants. The patterns' supports is mapped to a column on the matrix left side (\textbf{5}). The cumulative dataset coverage is mapped to a column on the matrix left side (\textbf{6}), representing the cumulative percentage of instances in a dataset captured by the patterns considering the matrix order (top to bottom). The variable importance is outlined above the global histograms, with the variables' names at the bottom (\textbf{7}). Pattern support, cumulative coverage, and variable importance are mapped to brightness (linear grayscale) and size (rectangle width). The FET p-value for each pattern is displayed in a column to the matrix right side (\textbf{8}) using a binary color scheme, green for values below $0.05$ (statistically significant) and purple otherwise.

We explore two aspects to help create more meaningful matrix-based visual representations for JEPs: filtering and ordering. Filtering is a common strategy when several patterns are available~\cite{Vico:2018:Anoverview, Loyola:2020:AReview}. JEPs can be filtered by support, data coverage, class, and supported instance(s) in our approach. Rows' and columns' order also plays an essential role in matrix visualization~\cite{Chen:2004:MatrixVisualization, Wu:2008:MatrixVisualization}. Our implementation supports different ordering schemes to enhance users' analytical capabilities. JEPs (rows) can be ordered by support, class, and class \& support. Furthermore, variables (columns) can be ordered by importance, which is calculated based on~\cite{Paja:2018:ADecision} as follows

\begin{equation}
  Imp(v_i, P) = \stackrel[j=1]{|P|}{\sum} \begin{cases}
    Supp(X_{o}, p_{j}) & \text{If $S_i \in p_{j}$} \\
    0 & \text{otherwise}
  \end{cases}
  \label{eq:IMP}
\end{equation}

where the importance of a variable $v_i$ given a set of JEPs $P$ is the summation of the support of each pattern $p \in P$ having a selector for $v_i$. After calculating the importance for all $v \in V$, they are normalized between $0$ and $1$.

\begin{figure}[h]
    \centering
    \includegraphics[width=.95\columnwidth]{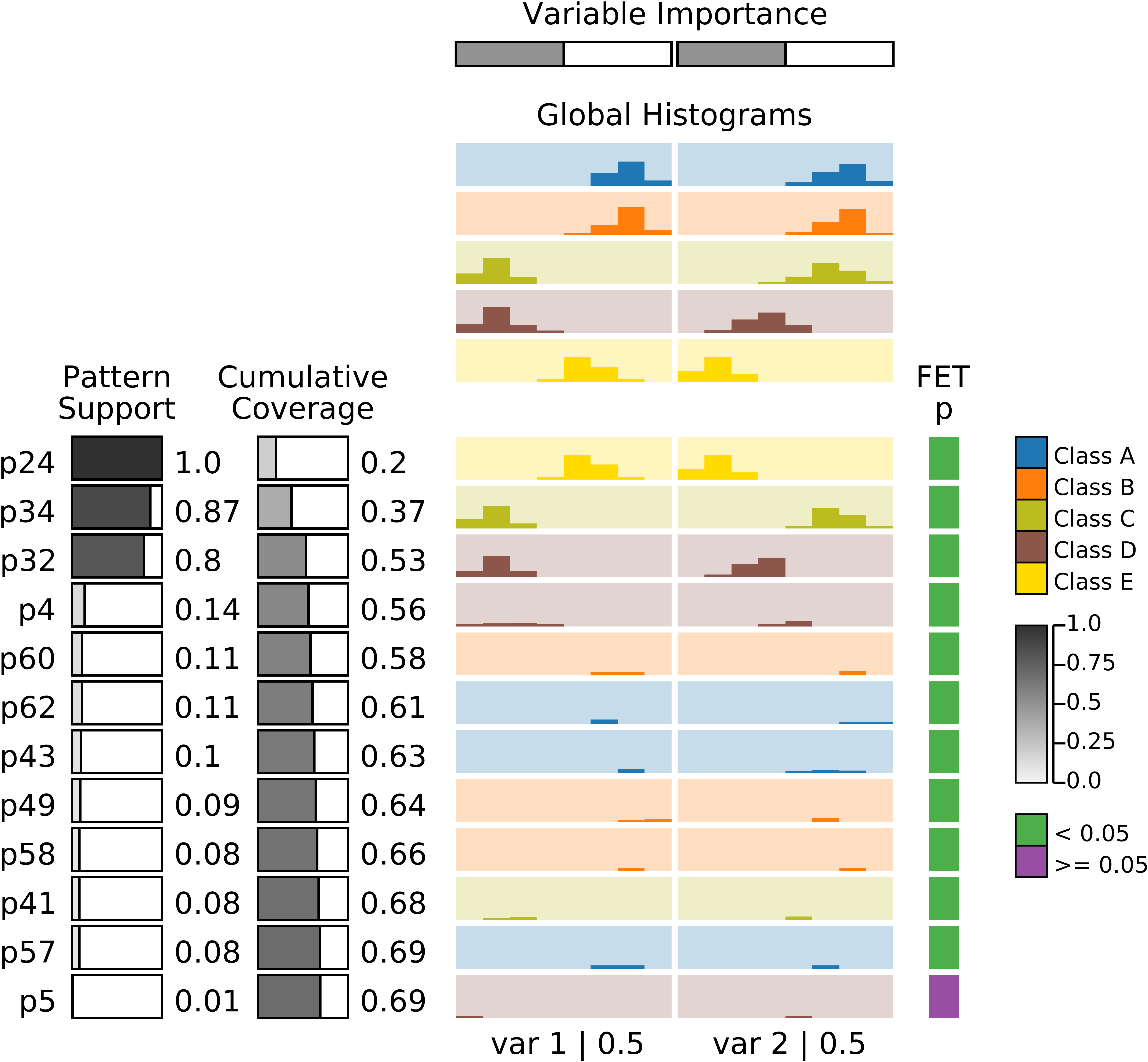}
    \caption{Twelve aggregated JEPs filtered out of the $67$ obtained for the synthetic dataset $X_{S}$. Variables values combinations and classes associations are presented, such as the strong pattern $p_{24}$, which supports all instances of class $E$ and indicates that these instances have median values for variable $var_{1}$ and low values for variable $var_{2}$. These $12$ JEPs represent about $69\%$ of the dataset $X_{S}$ (cumulative support), where only the low support pattern $p_{5}$ is not statistically significant (purple for FET p-value).}
    \label{fig:SD-EXP}
\end{figure}

Twelve JEPs are displayed in Fig.~\ref{fig:SD-EXP} for the synthetic dataset $X_{S}$ out of the $67$ previously discussed. The $67$ patterns were filtered by data coverage ($p_{24}$ to $p_{57}$) and supported instances ($p_{5}$). The $12$ resulting patterns are ordered by support. The first row contains the pattern $p_{24}$, where support is maximum, indicated by the filled rectangle on the support column. The cumulative coverage reflects the value of $0.20$ ($20\%$), as pattern $p_{24}$ supports all $100$ class $E$ instances out of $500$ dataset instances. Once pattern $p_{24}$ has selectors on variables $var_{1}$ and $var_{2}$, its cells show histograms for these variables considering only the instances (local) supported by $p_{24}$. For this case, global and local histograms are equal since $p_{24}$ supports all class $E$ instances. The second row represents the pattern $p_{34}$, with support equal to $0.87$, meaning that this pattern supports $87\%$ of class $C$ instances ($87$ instances out $100$). The cumulative coverage indicates the value of $0.37$, implying that patterns $p_{24}$ and $p_{34}$ together cover $37\%$ of the instances in $X_{S}$. The last pattern $p_{5}$ has significantly lower support ($0.01$), not being statistically significant (purple for FET p-value). All the remaining $11$ patterns are otherwise significant. From the first pattern $p_{24}$ to the last $p_{5}$, about $69\%$ of the dataset $X_{S}$ is represented (indicated by the cumulative coverage of $0.69$).

It is possible to observe in the JEPs of Fig.~\ref{fig:SD-EXP} variables values combinations associated with classes (insights \textbf{I1} and \textbf{I2}). Class $E$ is represented by the strong pattern $p_{24}$, unique and supporting $100\%$ of its instances (support value of $1.0$). Instances from this class have median values for the variable $var_{1}$ and low for the variable $var_{2}$, as represented by the local histograms. Patterns $p_{34}$ and $p_{32}$ complete the list of high support patterns. The $p_{34}$ supports $87\%$ of class $C$ instances, having low values for variable $var_{1}$ and high for variable $var_{2}$. Pattern $p_{32}$ supports $80\%$ of class $D$ instances, with low values for variable $var_{1}$ and median for variable $var_{2}$. These three patterns ($p_{24}$, $p_{34}$, and $p_{32}$) cover more than half ($53\%$, cumulative coverage of $0.53$) of the dataset. Pattern $p_{4}$ refers to a small subset of class $D$ instances (14\%), differing from pattern $p_{32}$ on variable $var_{2}$ (higher values). The patterns providing the highest support for classes $A$ and $B$ ($p_{60}$ and $p_{62}$) represent only $11\%$ of each class. Classes $A$ and $B$ strongly overlap (Fig.~\ref{fig:SD-SP}), so high support patterns for these classes are not attainable. Pattern $p_{41}$ represents a small subset of class $C$ instances ($8\%$), differing from pattern $p_{34}$ on variable $var_{2}$ (lower values). For class $D$, the pattern $p_{5}$ supports only one instance, representing an outlier compared to $p_{32}$ and $p_{4}$ on variable $var_{2}$ (higher value).

\subsection{Similarity Map}
\label{sec:map}

To support clusters and outliers analyses (insights \textbf{I3} and \textbf{I4}) from a class-labeled dataset $X$, we use DR layouts leveraging the space extension approach proposed in~\cite{Perez:2015:Interactive}. Our primary goal in using such an extension is to provide a straightforward way to navigate between patterns and data instances as ``viewed'' or represented by the patterns (insight \textbf{I5}), incorporating the JEPs' perspectives into the obtained layout. If only the original data is used, the pattern information is ignored. Using such an extension, it is easier to see the groups of instances that are well described by a pattern and the instances that are not. In general lines, a new variable is created for each data variable where the value of a given instance is calculated considering the pattern supporting it. Hence, instances supported by the same pattern receive the same value derived from the original variable. Formally, let $X \in \mathbb{R}^{N \times M}$ be the dataset, where $N = |X|$ is the number of instances in $X$ and $M = |V|$ the number of variables, the key idea is to create an extended dataset  $X' \in \mathbb{R}^{N \times 2M}$ as

\begin{equation}
    X' = [X|\widetilde{X}]
\end{equation}

with $\widetilde{X} \in \mathbb{R}^{N \times M}$ composed by the centroids (mean values) of disjoint groups of instances. For the instances subset $X_{h} \subset X$ belonging to a group $h$, the extension equals to

\begin{equation}
    \widetilde{x} = \frac{1}{|X_{h}|} \sum_{x \; \in \; X_{h}} x
\end{equation}

Since each instance $x_i \in X$ is supported by only one aggregated JEP, and we want to represent the relationships between instances from the patterns' perspective (insight \textbf{I5}), we use JEPs to define the groups of instances to extend $X$. Hence, $X'$ incorporates a new set of variables $\widetilde{V}$, representing the mean of the instances subset supported by each pattern.

To help users control the influence of the extension into the similarity relationships, a real parameter $\lambda \in [0,1]$ is used to control the gradual transition~\cite{Perez:2015:Interactive} between the dataset $X$ and the extended part $\widetilde{X}$ with $X_{weight} = X'W_{\lambda}$, where matrix $W_{\lambda} \in \mathbb{R}^{2M \times 2M}$ is defined by

\begin{equation}
    W_{\lambda} = 
    \left(
        \begin{array}{cc}
            (1 - \lambda)I & 0 \\
            0 & \lambda I
        \end{array}
    \right)
\end{equation}

After the normalization of $X_{weight}$ (z-score), a DR technique is used to create the similarity map. By varying the parameter $\lambda$, it is possible to create maps representing the distance relationships considering only the original dataset $X$ ($\lambda = 0$), only the extended part  $\widetilde{X}$ ($\lambda = 1$), or in between ($0 < \lambda < 1$), defining the similarity among instances using a combination of data and patterns information (in Sec.~\ref{subsec:usecasei} we present a heuristic to determine $\lambda$). In this paper, we use the Multidimensional Scaling (MDS) technique~\cite{Kruskal:1964:MDS} to project $X_{weight}$ since it faithfully preserves the global relationships between groups of instances. Other faster global techniques could also be used, for instance, the Landmarks MDS~\cite{lmds2004} to speed up this process, especially for larger datasets.

\begin{figure}[h]
    \centering
    \includegraphics[width=.9\columnwidth]{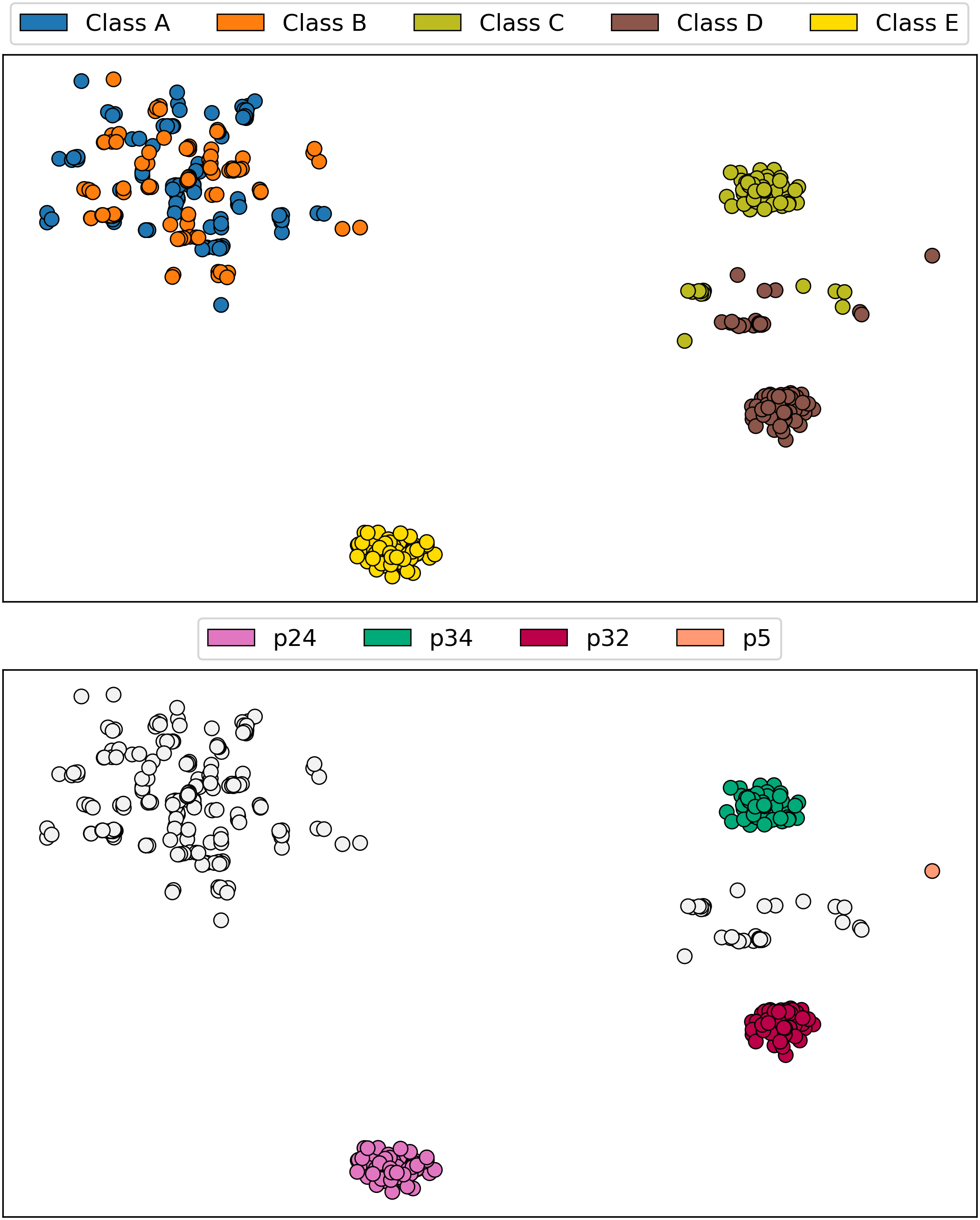}
    \caption{Similarity Map for the synthetic dataset $X_{S}$ considering the JEPs perspectives. Clusters can be spotted, formed by high support patterns $p_{24}$ (pink), $p_{34}$ (emerald), and $p_{32}$ (red). A class $D$ outlier can also be seen, represented by pattern $p_{5}$ (peach).}
    \label{fig:SD-MDS-L0_70}
\end{figure}

For creating the extended version of the synthetic dataset $X_{S}$, the $67$ JEPs previously discussed are used; that is, $67$ groups are considered in the extension process~\cite{Perez:2015:Interactive}. The result synthetic dataset extension $X'_{S}$ has then four variables $V'_{S} = \{ var_{1}, var_{2}, v\widetilde{ar}_{1}, v\widetilde{ar}_{2} \}$, where $v\widetilde{ar}_{1}$ and $v\widetilde{ar}_{2}$ are the mean values of $var_{1}$ and $var_{2}$ from each instances subset supported by the $67$ patterns. Fig.~\ref{fig:SD-MDS-L0_70} presents instances maps for $X_{weight} = X'_{S}W_{\lambda}$ with $\lambda = 0.70$. Two versions are present, one with colors given by the original dataset classes and another colored according to some JEPs (with gray indicating instances not supported by any pattern of interest). Three highly dense clusters can be identified (insight \textbf{I3}), one for each high supported pattern ($p_{24}$, $p_{34}$, and $p_{32}$). By inspecting the JEPs matrix (Fig.~\ref{fig:SD-EXP}), it is possible to observe (insight \textbf{I5}) that the one formed by pattern $p_{24}$ (insight \textbf{I1}) for class $E$ (pink) contains instances with high values for variable $var_{1}$ and low for variable $var_{2}$ (insight \textbf{I2}) compared to those established by patterns $p_{34}$ (emerald) and $p_{32}$ (red). The difference between these two latter clusters is found in variable $var_{2}$ (insight \textbf{I2}), where class $C$ instances ($p_{34}$) have higher values than class $D$ instances ($p_{32}$). Furthermore, an outlier for class $D$ (peach) can also be noticed  (insight \textbf{I4}). It resulted from the low support pattern $p_5$ (insight \textbf{I1}), with a low value for variable $var_{1}$ and a high for variable $var_{2}$ (insight \textbf{I2}) in contrast (exception) to the class $D$ cluster (red) settled by pattern $p_{32}$ (insight \textbf{I5}).

Similarity maps can be used to filter the JEPs matrix visualization, allowing users to select instances of interest, and obtaining the patterns supporting them. Similarity maps can play an essential role when dealing with a high number of patterns.

\section{Use Cases}
\label{sec:results}

This section presents two use cases showing how to use VAX for multivariate exploratory analysis. The two use cases involve datasets with and without ground truth labels (classes). VAX is implemented using Python and is also available as a code package~\footnote{\scriptsize \url{https://pypi.org/project/vaxm/}}\footnote{\scriptsize \url{https://popolinneto.gitlab.io/vaxm/papers/2022/tvcg/methodology/}}. The source code for the two use cases is accessible as Python notebook pages~\footnote{\scriptsize \url{https://popolinneto.gitlab.io/vaxm/papers/2022/tvcg/usecasei/}}\footnote{\scriptsize \url{https://popolinneto.gitlab.io/vaxm/papers/2022/tvcg/usecaseii/}}.

\subsection{Use Case I -- US Presidential Election}
\label{subsec:usecasei}

The first use case involves the analysis of the \textit{2016 US presidential election} using the dataset from a survey conducted in 2018 by the independent social research organization NORC of the University of Chicago~\cite{Tompson:2018:AP, Knittel:2020:Visual}. It contains the participants' answers about different political and societal aspects of the United States and what candidate they supported for the election (Donald Trump or Hillary Clinton). Following the steps described in~\cite{Knittel:2020:Visual}, the nationally representative subset is used, resulting in $4,913$ data instances (registered voters) and $67$ variables. After removing missing values and keeping only data instances with a revealed vote (Donald Trump or Hillary Clinton), the number of variables is reduced to $60$ and instances to $3,754$, $43.3\%$ pro-Donald Trump ($1,625$), and $56.7\%$ pro-Hillary Clinton ($2,129$).

\begin{figure}[!h]
    \centering
    \subfigure[Dataset coverage vs. Number of trees ($k$). With $2,048$ trees, data coverage is $100\%$.]
    {\includegraphics[width=\linewidth]{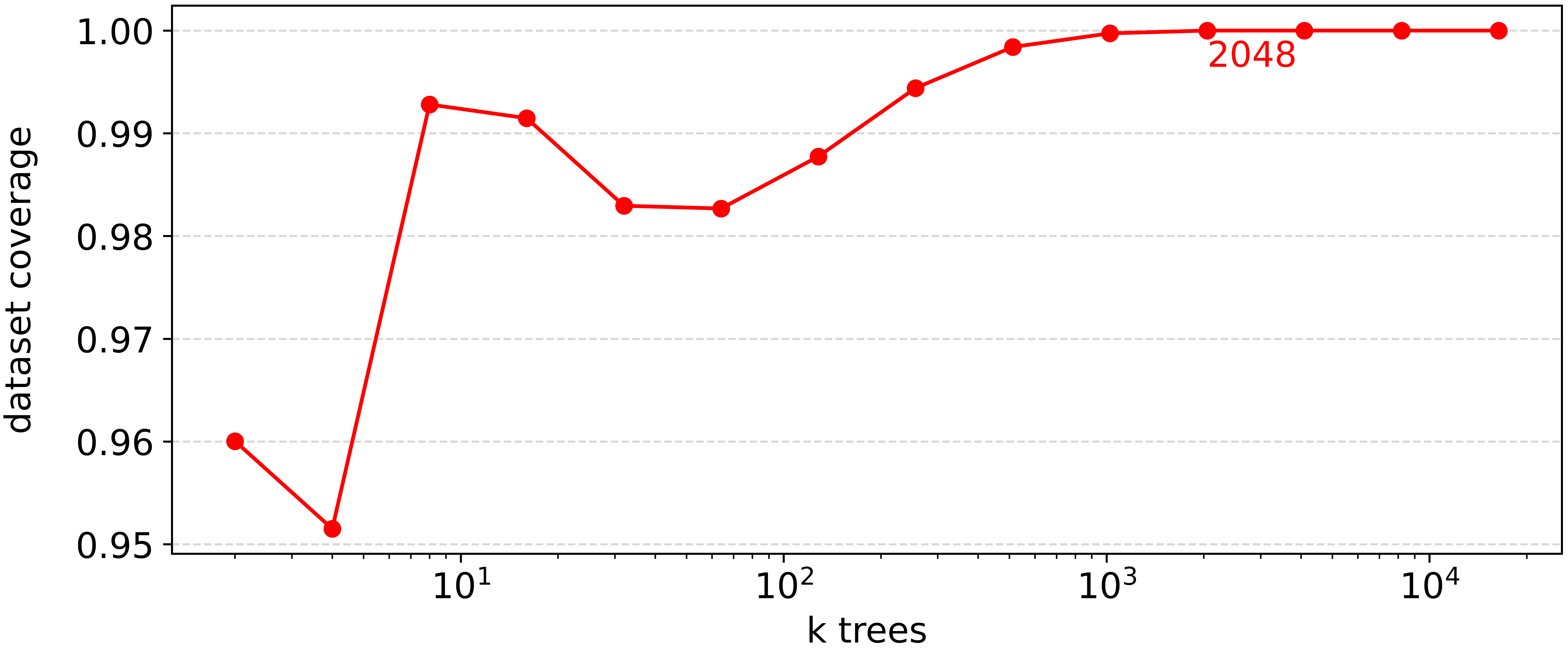}
    \label{fig:KTreesDataCoverage}}
    \subfigure[Kruskal Stress and Inverted Silhouette Coefficient of dataset extensions varying $\lambda$. The best tradeoff that minimizes stress (increase distance preservation) and silhouette (increase group cohesion/separation) is between $0.60$ and $0.70$.]
    {\includegraphics[width=\linewidth]{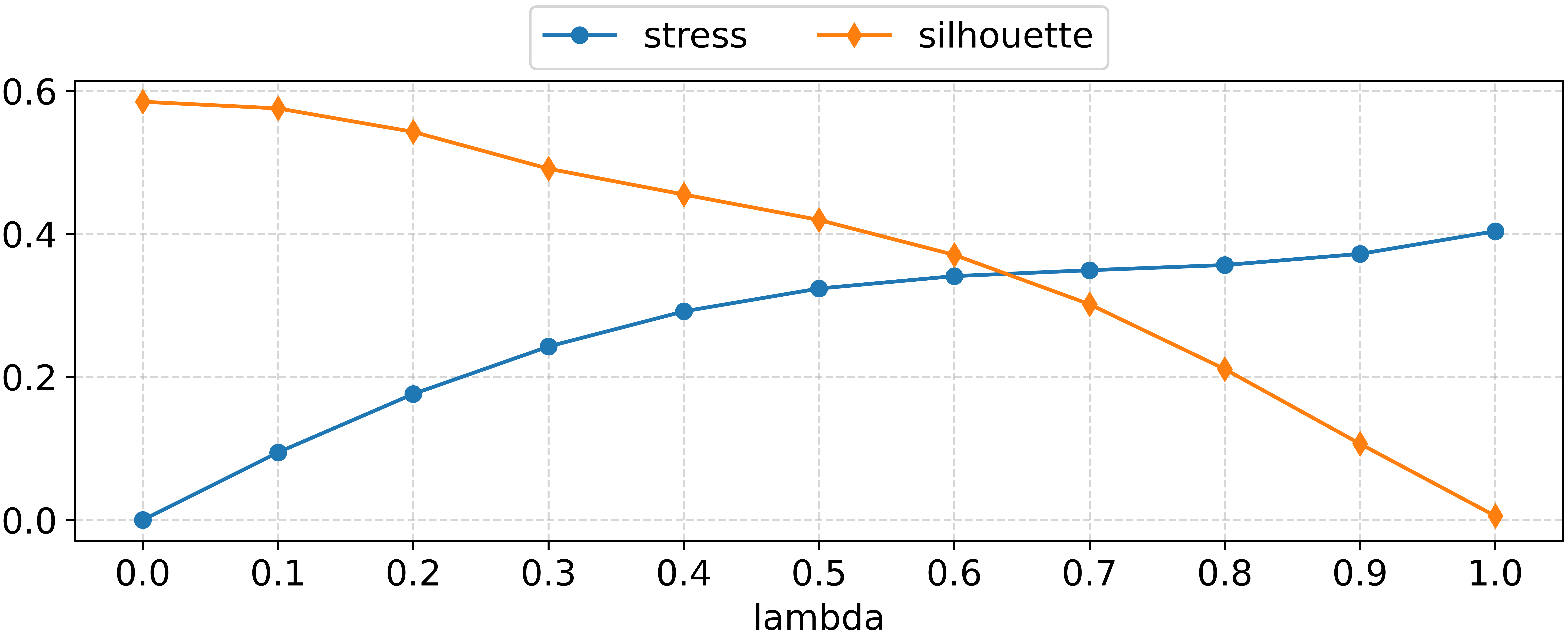}
    \label{fig:LambdaStressSilhouette}}
    \caption{Parameters analysis to set the number of trees $k$ for Algorithm~\ref{alg:EXT} and $\lambda$ for dataset extension considering the 2016 US presidential election dataset.}
    \label{fig:PAR}
\end{figure}

The first step in the VAX pipeline is to induct the random DTs to extract the patterns (Algorithm~\ref{alg:EXT}). In this process, the only parameter that needs to be set is the number of trees. To reduce pattern duplication, we use the minimum number of trees that result in patterns representing all data instances (after Algorithm~\ref{alg:SELAGG}), starting with a small number of DTs and successively inducting more trees until the data coverage is $100\%$. Fig.~\ref{fig:KTreesDataCoverage} presents the obtained data coverage for different numbers of trees, from $2^{1}=2$ to $2^{14}=16,384$. In this experiment, $2,048$ is the minimum number of DTs capable of providing enough diversified patterns resulting in $100\%$ of coverage after selection and aggregation. After extraction, from the $856,900$ discovered patterns, $255$ are selected, $3,042$ aggregated, and $853,603$ discarded (Algorithm~\ref{alg:SELAGG}).

For the similarity map creation, $\lambda$ must be defined for the dataset extension~\cite{Perez:2015:Interactive}. In this process, we propose to select $\lambda$ by analyzing the Kruskal Stress~\cite{Kruskal:1964:MDS} and Silhouette Coefficient~\cite{Rousseeuw:1987:Silhouettes}. The idea is to find the value that maximizes distance preservation (stress) and group cohesion/separation (silhouette) so that groups defined by patterns can be spotted while minimizing the distortion of distances conveyed by the original data. To make our analysis more straightforward, we normalize and invert the original Silhouette Coefficient, using $sc' = 1 - \frac{sc+1}{2}$, where $sc$ is the original silhouette, resulting in values in $sc' \in [0, 1]$ (the lower, the better). Fig.~\ref{fig:LambdaStressSilhouette} shows the Kruskal Stress $st \in [0,1]$ (the lower, the better) and the inverted Silhouette Coefficient $sc'$ of different extended datasets (using the $255$ selected and aggregaed patterns) varying $\lambda$. The best tradeoff between distance preservation and group cohesion/separation is found between $0.60$ and $0.70$. Hence, we empirically set $\lambda = 0.65$. Notice that $\lambda$ is calculated after the logic rules inference, selection, and aggregation, so this process involves the projection of the extended datasets and the stress and silhouette computation only.

\begin{figure*}[h]
    \centering    
    \subfigure[Fourteen JEPs filtered out the $255$ selected and aggregated. Rows (JEPs) are ordered by support and columns (variables) by importance.]
    {\includegraphics[width=\linewidth]{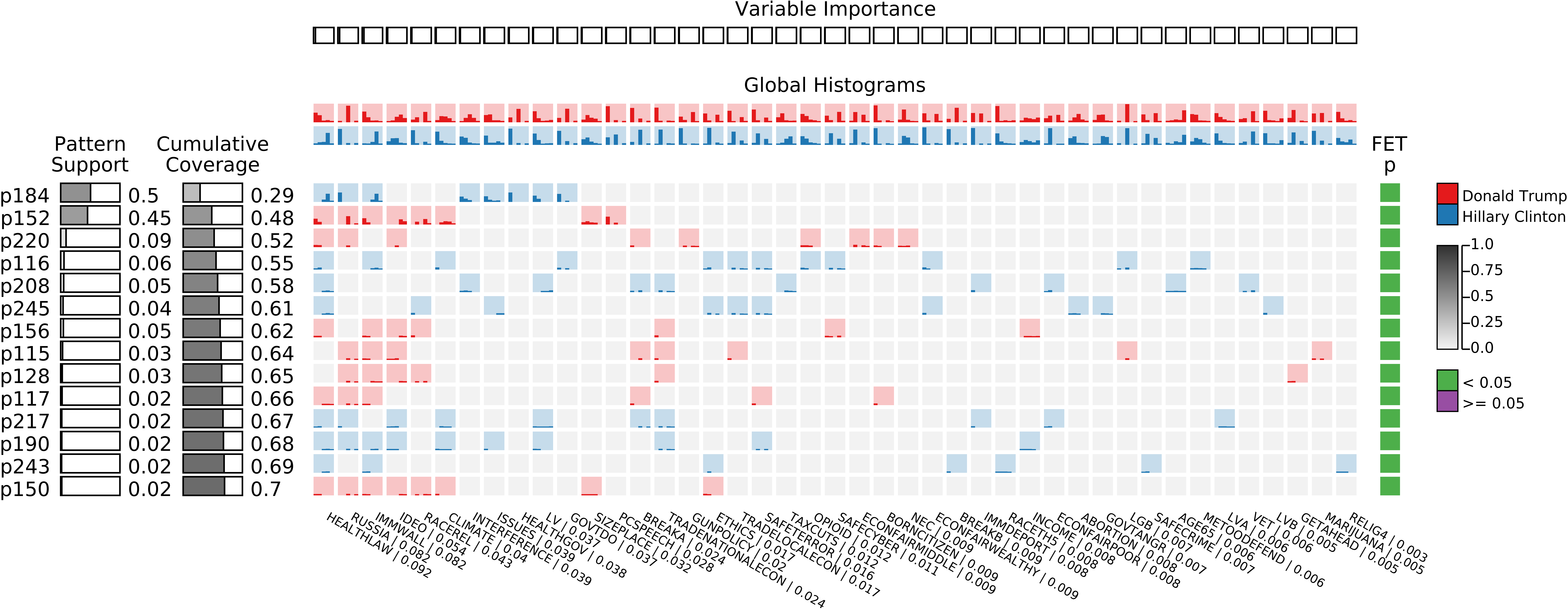}\label{fig:EL-EXP}}    
    \subfigure[The similarity map for $\lambda = 0.0$.]
    {\includegraphics[width=0.35\linewidth]{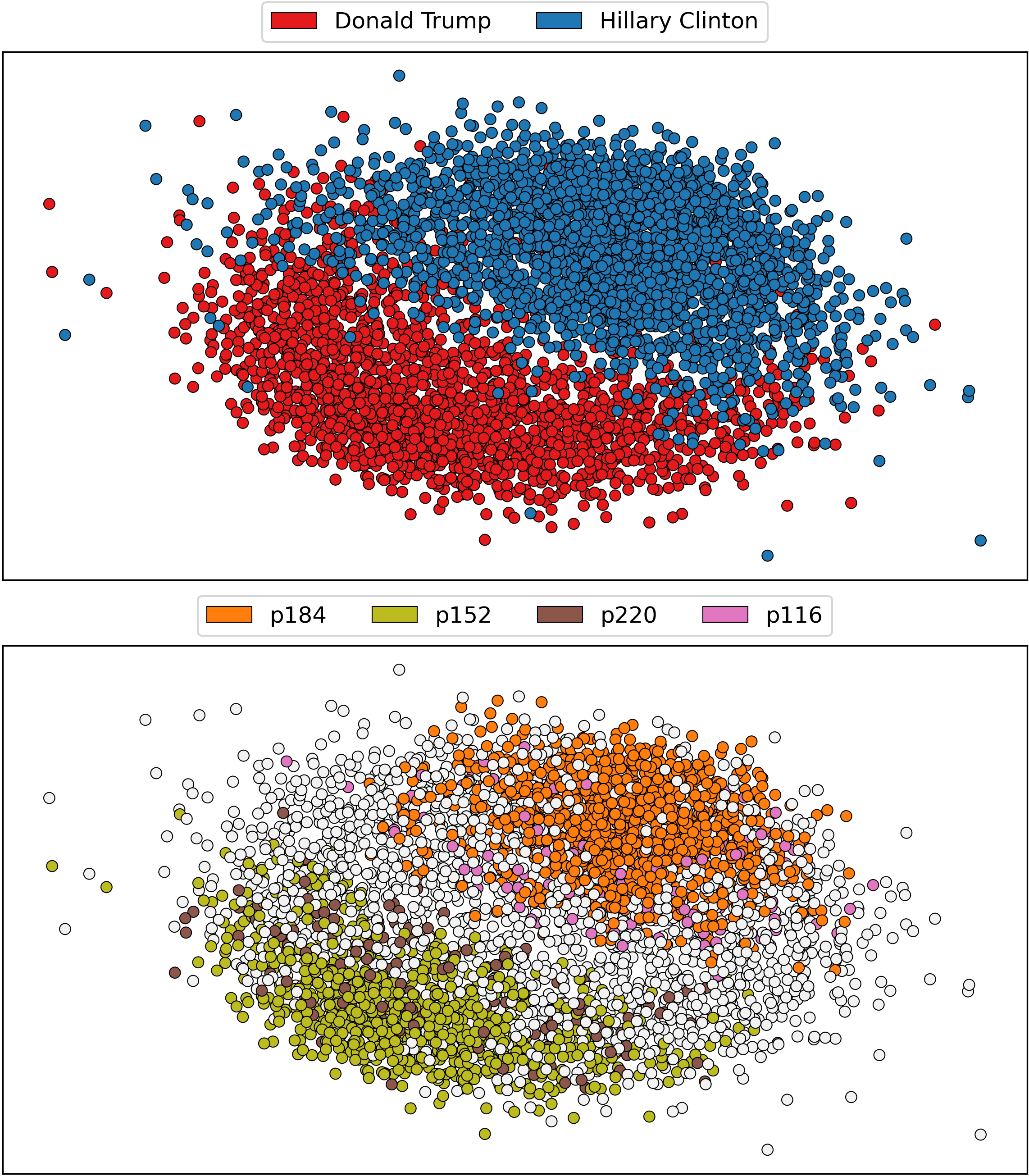}\label{fig:EL-MDS-L0_00}}\quad
    \subfigure[The similarity map for $\lambda = 0.65$.]
    {\includegraphics[width=0.35\linewidth]{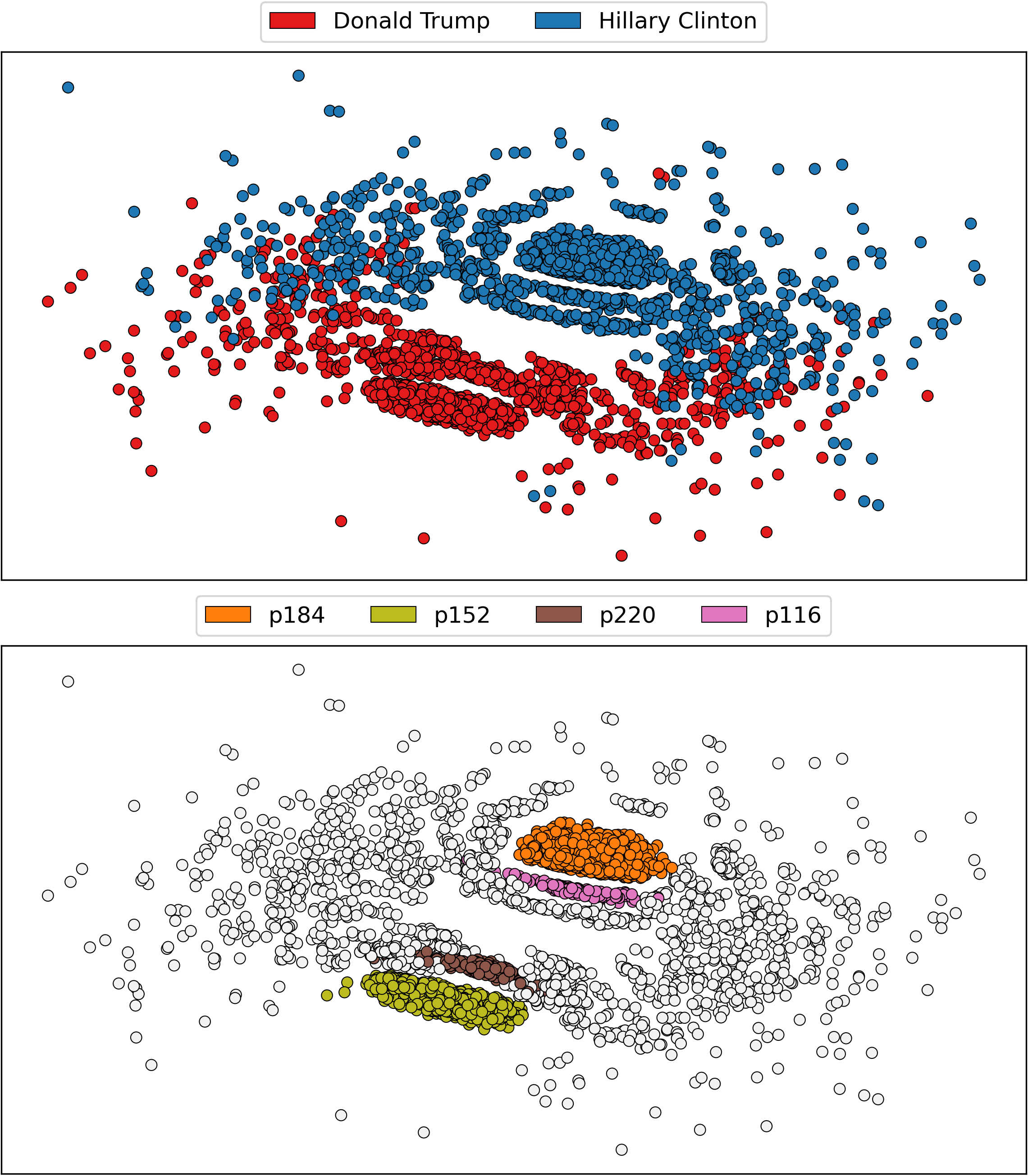}\label{fig:EL-MDS-L0_65}}\\    
    \caption{The JEPs matrix and similarity maps for the 2016 US election dataset. By inspecting JEPs (a), about half ($48\%$) of the electorate can be described by only two patterns ($p_{184}$ and $p_{152}$), and these diverge in three points. The support for the Affordable Care Act, the construction of the wall with Mexico, and the Russian participation in Trump's campaign. The $4$ highest support patterns ($p_{184}$, $p_{152}$, $p_{220}$, and $p_{116}$) can be seen as clusters in the similarity map (c). Notice we use a linear scale to represent the variable importance in our examples. A log scale could also be used, helping to make clear the differences when values are small.}    
    \label{fig:EL}
\end{figure*}

Fig.~\ref{fig:EL-EXP} presents $14$ JEPs from the resulting $255$, representing $70\%$ of the dataset. The patterns are ordered by support and variables by importance. Despite the number of variables and complexity of the dataset, about half of its instances ($48\%$) are described by only two patterns, $p_{184}$ and $p_{152}$ (first two rows). These are the highest support patterns for Hillary and Trump voters, respectively. Interestingly, it is possible to notice, given the differences in the local histograms of both patterns, that half of the Hillary voters ($50\%$ from pattern $p_{184}$) and about half of Trump voters ($45\%$ from pattern $p_{152}$) differ fundamentally in three aspects (first three columns). While Hillary voters are in favor of expanding the Affordable Care Act -- Obamacare (HEALTHLAW variable with majority answers ``Expand the law''), against the wall with Mexico (IMMWALL variable with majority answers ``Strongly oppose''), and believe that the Trump election campaign was coordinated with Russia (RUSSIA variable with majority answers ``Yes''), Trump voters have opposite opinions (HEALTHLAW variable with majority answers ``Repeal the law entirely'', IMMWALL variable with majority answers ``Strongly favor'', and ``RUSSIA'' with majority answers ``No''). Notice that, although these three points compose the more generic differences between almost half of the voters, the differences are not evident for the other half, which can be observed on the global histograms for these three variables.

The support drops considerably for the subsequent patterns. However, revealing a more heterogeneous scenario for the other half of the voters. For instance, pattern $p_{220}$ (third row) diverges from the strong pattern $p_{152}$ (second row), both representing Trump voters, on political ideology (variable IDEO fourth column) with voters described by  $p_{220}$ considering themselves as moderate instead of conservative.  Pattern $p_{116}$ (fourth row) differs from the strong pattern $p_{184}$ (first row), both describing Hillary voters, regarding the Affordable Care Act (variable HEALTHLAW first column), advocating to repeal the law at least in parts. The last pattern $p_{150}$ supports only $2\%$ of instances from its associated class. All the remaining $241$ patterns have equal or less than $2\%$ of support.

In addition to the JEPs matrix, two similarity maps are created to help the analysis, one in Fig.~\ref{fig:EL-MDS-L0_00} considering the original dataset ($\lambda = 0$) and another in Fig.~\ref{fig:EL-MDS-L0_65} using an extended version ($\lambda = 0.65$). In both figures, the maps at the top are colored using voting information (Donald Trump or Hillary Clinton), and the maps at the bottom using pattern information, considering the $4$ patterns with the highest support ($p_{184}$, $p_{152}$, $p_{220}$, and $p_{116}$). Instances colored in gray represent voters not supported by these $4$ patterns. These high support patterns compose clusters in the similarity maps of Fig.~\ref{fig:EL-MDS-L0_65} but not of Fig.~\ref{fig:EL-MDS-L0_00}, indicating that adding JEPs information can reveal clusters in DR layouts that are otherwise not found using only the original dataset, which helps users navigate between patterns and data.

Finally, while the majority of the US electorate can be split into two major groups, one supporting Hilary and another Trump, a fascinating picture emerge. The usual simplistic division between two extremes is indeed much more complex in real life, with several subgroups diverging from the most general ``common sense'' within a group. This is clear in the similarity map where the major groups are spotted, but different patterns conveyed by the JEPs also define mixed subgroups.

\subsection{Use Case II -- World Happiness}

\begin{figure*}[h]
    \centering    
    \subfigure[Six filtered JEPs out of the $29$  selected and aggregated. Matrix rows (JEPs) are ordered by support and columns (variables) by importance.]
    {\includegraphics[width=0.65\linewidth]{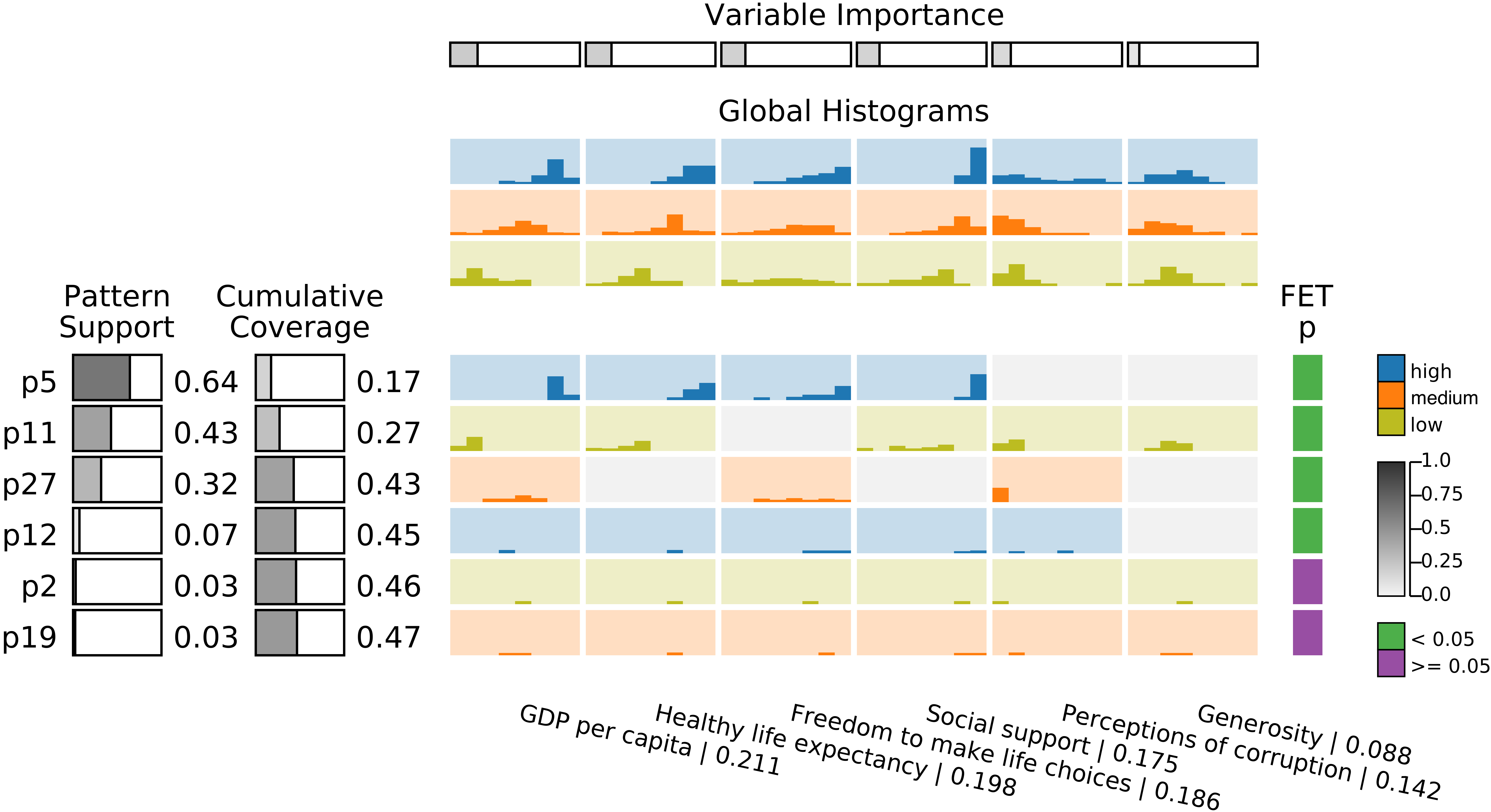}\label{fig:WH3C-EXP}}\quad
    \subfigure[The similarity maps with $\lambda = 0.65$.]
    {\includegraphics[width=0.30\linewidth]{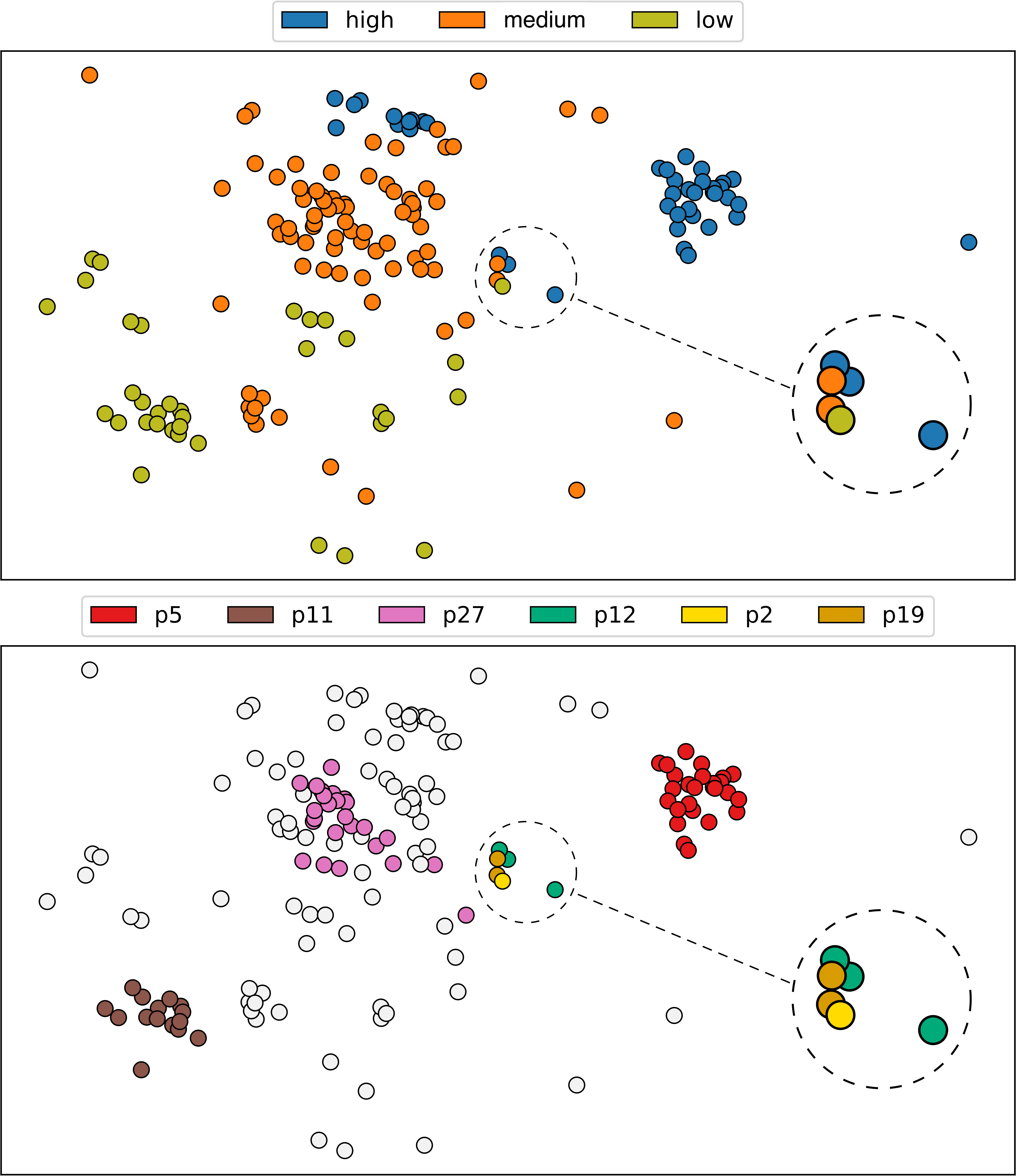}\label{fig:WH3C-MDS-L0_65}}\\    
    \caption{The JEPs matrix and similarity maps for the 2019 World Happiness Report dataset. The first $3$ patterns ($p_{5}$, $p_{11}$, and $p_{27}$) in (a) describe the general behavior of high, medium, and low happy countries, with a clear difference between them in terms of ``GDP per capita'', ```Healthy life expectancy'', and ``Social support''. The similarity maps (b) allow reasoning about clusters and outliers. The $6$ countries located at the center of the map are selected for further analysis.}    
    \label{fig:WH3C}
\end{figure*}

The second use case involves the dataset of the \textit{2019 World Happiness Report} made available by the \textit{Sustainable Development Solutions Network (SDSN)}~\cite{Helliwell:2019:WH19}. This dataset presents the ``Happiness Score'' of $156$ countries, indicating how happy their citizens perceive themselves, and six other variables capturing different perspectives. These variables are ``GDP per capita'', ``Social support'', ``Healthy life expectancy'', ``Freedom to make life choices'', ``Generosity'', and ``Perceptions of corruption''.

Since the world happiness dataset does not present class labels, we discretize the happiness score into three groups, indicating low, medium, and high happy countries. The idea is to verify the differences between countries based on perceived happiness levels regarding the other six variables. The happiness score is discretized in three equally sized bins. In other words, we are transforming a regression problem into a classification~\cite{Salman:2012:Regression}. We have chosen discretization, but approaches such as clustering~\cite{James:2013:An} and instances subset selection~\cite{Cao:2020:DRIL} are also suitable.

The ``high happy'' class encloses $42$ countries with happiness scores in $[6.13, 7.76]$, the ``medium happy'' contains $79$ countries with scores in $[4.49, 6.13]$, and the ``low happy'' class groups $35$ countries with scores in $[2.85, 4.49]$. Since the variable ``Happiness Score'' is employed to define the classes (discretization), it is not used to create the patterns. The number of trees $k$ and $\lambda$ are set following the procedures specified in Sec.~\ref{subsec:usecasei}. From $2,893$ extracted patterns ($k = 64$), $29$ were selected, $252$ aggregated, and $2,612$ discarded. For the similarity map creation, we use $\lambda = 0.65$.

Out of the $29$ JEPs, $6$ were selected to create the JEPs matrix of Fig.~\ref{fig:WH3C-EXP}, one per class with the highest supports per class and $3$ others representing the highlighted instances in Fig.~\ref{fig:WH3C-MDS-L0_65}. These $6$ JEPs represent $47\%$ of the dataset. The patterns are ordered by support and variables by importance in the resulting matrix. Pattern $p_{5}$ represents $64\%$ of the ``high happy'' countries, indicating that those countries have high values in ``GDP per capita'', ``Health life expectancy'', and ``Social support''. Moreover, $p_{5}$ supports countries with varied values for ``Freedom to make life choices''. Pattern $p_{11}$ describes ``low happy'' countries, accounting for $43\%$ of that class. These countries have low values for ``GDP per capita'' and ``Perceptions of corruption'' and median values for ``Health life expectancy'', ``Social support'', and ``Generosity''. Compared to ``high happy'' countries (pattern $p_{5}$), it is clear that those ``low happy'' countries present lower values for ``GDP per capita'', ``Health life expectancy'', and ``Social support''. Pattern $p_{27}$ supports $32\%$ of ``medium happy'' countries, having median values for ``GDP per capita'', low values for ``Perceptions of corruption'', and flat distributed values for ``Freedom to make life choices''. These ``medium happy'' countries are between ``high happy'' and ``low happy'' in terms of ``GDP per capita'', but the other two employed variables, ``Freedom to make life choices'' and ``Perceptions of corruption'', assume values not much different from the other two classes.

The first three patterns $p_{5}$, $p_{11}$, and $p_{27}$ of Fig.~\ref{fig:WH3C-EXP} encode countries' general behavior ($43\%$ of the dataset), following the tendencies of the global histograms. The patterns $p_{5}$ and $p_{11}$ result in two clusters in the similarity map presented in Fig.~\ref{fig:WH3C-MDS-L0_65}. The map on the top presents instances colored by class (``high'', ``medium'', and ``low''), and the bottom by supported pattern ($p_{5}$, $p_{11}$, $p_{27}$, $p_{12}$, $p_{2}$, and $p_{19}$). The map allows users to reason about clusters and outliers by selecting instances of interest and visualizing their respective patterns. From Fig.~\ref{fig:WH3C-MDS-L0_65}, the $6$ countries placed at the center of the map are selected, presenting their patterns ($p_{12}$, $p_{2}$, and $p_{19}$) in Fig.~\ref{fig:WH3C-EXP} together with the other $3$ high supported patterns ($p_{5}$, $p_{11}$, and $p_{27}$). The $3$ ``high happy'' countries supported by pattern $p_{12}$ (emerald) differ from the general group of high happy countries ($p_{5}$) by having median values for ``GDP per capita''. The single ``low happy'' country supported by pattern $p_{2}$ (yellow) diverges from the general group of low happy countries ($p_{11}$) by holding a higher value of ``GDP per capita'', ``Healthy life expectancy'', and ``Social support''. The $2$ ``medium happy'' countries supported by pattern $p_{19}$ (gold) are different from the general group of medium happy countries ($p_{27}$) by yielding high values for ``Perceptions of corruption''.  Furthermore, these $6$ countries with different happiness levels are similar in ``Healthy life expectancy''. In summary, browsing the instances similarity map and visualizing patterns from selected instances makes it possible to reason about countries' differences and similarities under the optics of happiness levels, allowing the analysis of the most general behavior per class and the particularities of sub-groups or outliers that deviate from the norm.

\begin{figure}[h]
    \centering
    \includegraphics[width=\linewidth]{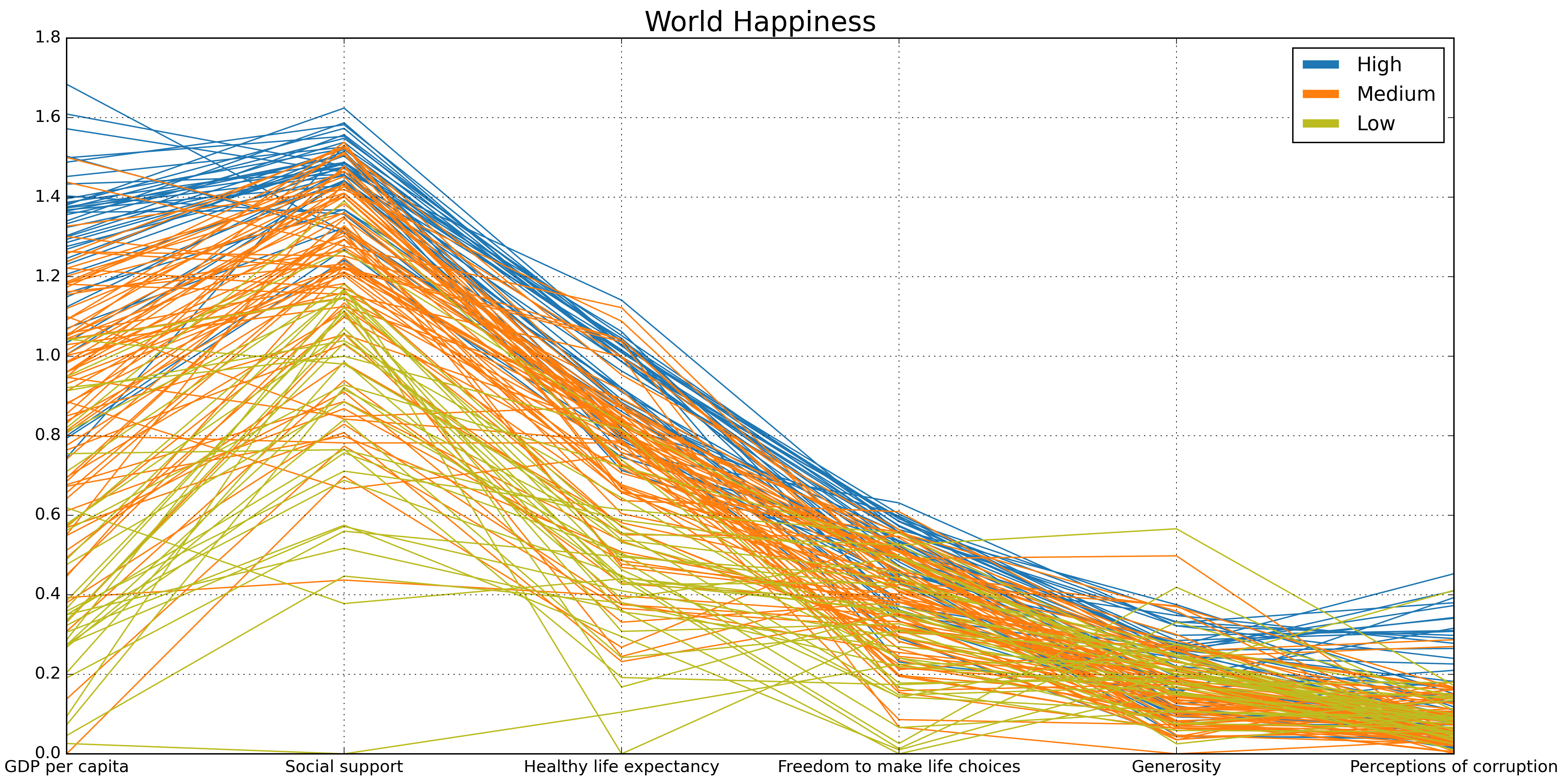}
    \caption{World Happiness Report PCP. General trends/patterns are possible to spot. However, quantifying the number of countries for which the general patterns are valid, including the range variables where they hold, is challenging. Also, finding representative (not general) intricate patterns can be, if possible, complex and time-consuming in practice.}
    \label{fig:woldhappinesspcp}
\end{figure}

This explicit discovery and representation of patterns enables VAX to help users uncover intricate relationships that are hard to capture by employing usual multidimensional visualization techniques. As an example, a Parallel Coordinates Plot (PCP) for the 2019 World Happiness Report dataset is presented in Fig.~\ref{fig:woldhappinesspcp}. In this visualization, it is possible to see general trends/patterns, conveying similar information as the patterns $p_{5}$, $p_{11}$, and $p_{27}$ (Fig.~\ref{fig:WH3C-EXP}). However, quantifying the number of countries for which this information is true, including the ranges of the variables where this holds, is challenging. The analysis becomes even more troublesome when representative secondary patterns exist but deviate from the general norm, e.g., the pattern $p_{12}$ (Fig.~\ref{fig:WH3C-EXP}). This pattern is valid for $7\%$ of highly happy countries and involves several different variable ranges. This intricate information, explicitly represented by the VAX pattern, is very hard, if possible, to be discovered using a PCP. Selection (interaction) may help to mitigate the problem but finding such patterns may be complex and time-consuming in practice.

Similar observations can be made about Scatterplot Matrices (SPLOMs). An intuition about general patterns can be acquired, especially about the individual variables' distributions. However, it is difficult to quantify the number of countries for which this information is valid and the range of variables and distribution where this holds -- in the accompanying supplemental material, we present such SPLOM in addition to the PCP and SPLOM visualizations of the \textit{US Election} dataset to emphasize the advantages of explicitly capturing and representing data patterns using VAX opposed to letting users find them using typical multidimensional visual representations.

\section{Discussion and Limitations}
\label{sec:limitations}

\noindent\textbf{Prediction vs. description.} The distinction between model interpretability (focus on prediction) and data explanation (focus on description) is blurred. At first glance, any combination of a prediction model and interpretability strategy could also be used for data description. While it is true that some (global) interpretability techniques can be adapted, a crucial difference must be observed for the general case. While prediction focuses on finding the best way to separate classes with generalization in mind so unknown instances can be properly classified, description (through JEPs) focuses on specialization since the goal is to understand all data of a given dataset, with overfitting an acceptable and positive property~\cite{Gleicher:2013:Expaliners,Loyola:2019:Fusing,Knittel:2020:Visual}. The result is that the inter-class separation is usually well represented by interpretability strategies. Still, intra-class information is typically lost, such as the existence of subgroups of instances within a class.

\vspace{0.25cm}\noindent\textbf{Existing solutions.} EPs visualization has great potential~\cite{Novak:2009:Supervised, Loyola:2020:AReview}, and VAX provides a powerful method to visualize JEPs focusing on multivariate data analysis via automated insights~\cite{Law:2020:Characterizing}. VAX stands out from the existing data explanation approaches, such as Explainers~\cite{Gleicher:2013:Expaliners} and Visual Neural Decomposition (VND)~\cite{Knittel:2020:Visual}, by allowing patterns with more than three variables, non-linear relations, and class ``pure'' patterns. Nevertheless, such patterns may have lower support when compared with patterns supporting instances from different classes (as found in VND~\cite{Knittel:2020:Visual}). If desirable, VAX is readily adaptable to display different types of EPs (not only on JEPs), mining patterns where instances belong to multiple classes as well. VAX differs from model explanations solutions, such as ExMatrix~\cite{PopolinNeto:2020:ExMatrix} and RuleMatrix~\cite{Ming:2019:RuleMatrix} by conveying visual representations that use local histograms (instead of variables ranges) and DR layouts that represent patterns perspectives. Furthermore, by using a matrix-like visual metaphor, VAX allows for producing meaningful visual representations by filtering and ordering patterns (rows) and variables (columns) beyond what was explored in this paper.

\vspace{0.25cm}\noindent\textbf{Pattern extraction, selection, and aggregation.} VAX involves extracting, selecting, aggregating, and visualizing JEPs. The RFm method~\cite{Borroto:2015:Finding, Loyola:2020:AReview} is used for building random DT models (Algorithm~\ref{alg:EXT}), and it has been proved to be a valuable approach to mine diversified patterns~\cite{Borroto:2015:Finding}. The JEPs selection and aggregation process (Algorithm~\ref{alg:SELAGG}) is inspired by an existing approach~\cite{Borroto:2015:Finding}, reducing the initial set of extracted patterns. These are selected by their confidence and support and aggregated with other meaningful patterns. Those not fulfilling the specifications of being representative (high support) or an exception (for particular instances) are discarded. This way, complementary patterns are aggregated, not losing their information, whereas redundant patterns are ignored.

In this paper, we suggested a simple approach to define $k$ for the extraction process to discover enough patterns covering all data instances. This is indeed an expensive approach, and usually setting $k$ to a large number (proportional to the dataset size), the usual literature's approach~\cite{Borroto:2015:Finding, Loyola:2019:Fusing, Vico:2018:Anoverview}, typically brings good results. Another possibility to speed up our process is setting a data coverage threshold (smaller than $100\%$). In this case, the downside is that some instances (probably outliers) may not be supported by patterns.

\vspace{0.25cm}\noindent\textbf{DR layouts extension.} DR layouts are great for analyzing multidimensional data~\cite{Nonato:2019:Multidimensional, Liu:2017:Visualizing, Perez:2015:Interactive}. Still, the quality of a layout is directly related to the DR technique's efficiency in revealing clusters and outliers presented in the original space. To help in this process, we extend the original space~\cite{Perez:2015:Interactive} creating new data variables that add pattern information so that the resulting layout also reflects the patterns. The DR layout can be browsed for clusters and outliers, presenting the patterns supporting instances of interest. With JEPs matrix visualization and the similarity maps, compound facts~\cite{Law:2020:Characterizing} can be generated, enabling linked data insights involving descriptive patterns, clusters, and outliers. Compound facts are highly desirable, providing more nuanced insights about multivariate data~\cite{Law:2020:Characterizing}.

Although not a constrain, we used the statistical mean in this process of variable creation, as suggested in~\cite{Perez:2015:Interactive}, and the classic MDS technique~\cite{Kruskal:1964:MDS} for generating DR layouts. Other statistical measures or DR techniques could also be used. The study of how these choices impact the produced DR layouts is fascinating but out of our scope. We leave such investigation as future work.

\vspace{0.25cm}\noindent\textbf{Scalability.} Since histograms require a certain vertical (height) display space, it is impossible to visualize at the JEPs matrix a significant number of patterns at once. Potential solutions are selecting groups of interest, like the patterns that combined attain minimum dataset support (e.g., 70\%) or focusing on particular instances (e.g., outliers). The visual metaphor is also limited by the number of variables used by a single pattern regarding horizontal space for displaying variables. For this matter, filtering variables by importance may be an option.

\vspace{0.25cm}\noindent\textbf{JEP's quality and ambiguous data.} The quality of the extracted JEPs is related to DTs' ability to capture the differences between distinct classes. Although JEPs have proven to be capable of capturing diversified relations~\cite{Borroto:2015:Finding}, non-linear relationships are represented through local linear orthogonal approximations, and such approximations may result in too many low-support JEPs. In the worst-case scenario, one JEP per instance if the separation between classes is non-existent, defeating the purpose of generating descriptions for general patterns in the data -- for instance, Classes A (orange) and B (blue) in Fig~\ref{fig:SD-SP}. This issue derives from DTs' limitation of learning generic (complex) patterns or their nonexistence in the dataset in question, which is desirable since it is better not to create artifacts.

JEPs' quality is not related to the usual definition of models' performance. Since ``all'' data is used for training, the testing accuracy is irrelevant. However, the training accuracy is crucial. The generated random DTs have to present $100\%$ training accuracy; with less than that, misclassified instances will be wrongly explained and may impact the variables' ranges on the extracted patterns, generating spurious information. One solution would be to ignore misclassified instances, but they may be essential for the task at hand. In our case, we guarantee $100\%$ accuracy for every DT by removing ambiguous instances (instances with the same values but belonging to different classes) as part of a pre-processing stage and by overfitting the models using not pruned DTs -- in our implementation, if the randomly selected variables to construct a tree do not result in a $100\%$ accurate model, other variables are added until such accuracy is reached. The downside is the high number of resulting low support patterns and the lack of support for ambiguous instances. In any case, since JEPs are class-pure patterns, it is unwanted (or even impossible) to have instances of different classes supported by the same pattern. So, removing ambiguous instances avoids creating erroneous explanations.

\section{Conclusions}
\label{sec:conclusions}

This paper presents \textit{multiVariate dAta eXplanation (VAX)}, a new Visual Analytics (VA) approach for analyzing multivariate datasets. VAX employs aggregated \textit{Jumping Emerging Patterns (JEPs)} to capture intricate patterns in class-labeled datasets. A matrix-like visual metaphor is used for JEPs visualization, where patterns are rows, variables are columns, and data distribution conveyed using histograms are matrix cells. Based on matrix visualization, meaningful visual representations can be reached by filtering and ordering patterns (rows) and variables (columns). Furthermore, similarity maps produced by DR techniques aim to convey a better overall image of a dataset (e.g., clusters and outliers) using the JEPs lens. VAX can be applied to different domains, which can address phenomenon comprehension through knowledge acquisition, being a valuable tool for extracting data insights. We plan as future work to develop new approaches for filtering and ordering JEPs and variables to enhance VAX visual representations. Moreover, we intend to pursue new methods for diversified pattern extraction and comparisons involving different DR techniques and statistical measures for similarity map creation.


%

%

\ifCLASSOPTIONcompsoc
  \section*{Acknowledgments}
\else
  \section*{Acknowledgment}
\fi

The authors wish to thank the valuable comments and suggestions obtained from the reviewers, as well as the support received from the Qualification Program of the Federal Institute of S\~ao Paulo (IFSP).

\ifCLASSOPTIONcaptionsoff
  \newpage
\fi



%
%
%

\bibliographystyle{IEEEtran}
\bibliography{References.bib}

%

\begin{IEEEbiography}[{\includegraphics[width=1in,height=1.25in,clip,keepaspectratio]{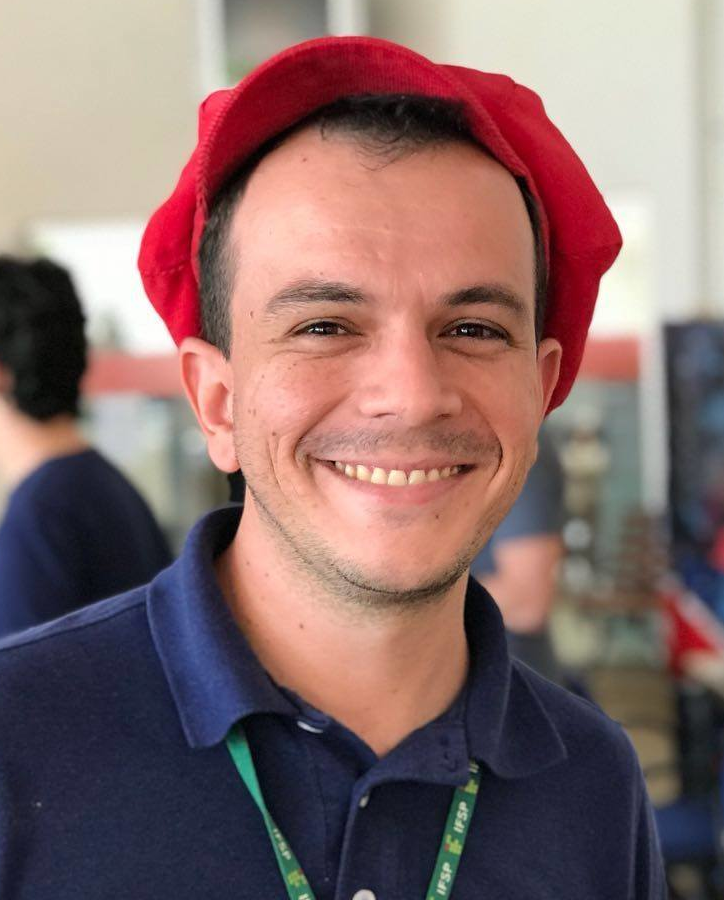}}]{M\'ario~Popolin~Neto}

received his Ph.D. degree in computer science from the University of S\~ao Paulo, Brazil, in 2021. Currently, he is a professor at the Federal Institute of S\~ao Paulo - IFSP, Araraquara - SP, Brazil. His research interests include Information Visualization and Visual Analytics. His work contributes towards visualization methods for classification models' interpretability, promoting visual explanations of models and data.
\end{IEEEbiography}

\begin{IEEEbiography}[{\includegraphics[width=1in,height=1.25in,clip,keepaspectratio]{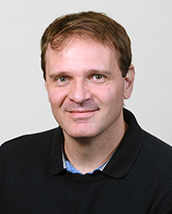}}]{Fernando~V.~Paulovich}
is an Associate Professor in Visual Analytics for Data Science at the Department of Mathematics and Computer Science, Eindhoven University of Technology (TU/e), the Netherlands. Before moving to the Netherlands, he was a Professor and Canada Research Chair at Dalhousie University, Canada (2017-2022), and an associate professor at the University of São Paulo, Brazil (2009-2017). He has been researching information visualization and visual analytics, focusing on integrating machine learning and visualization tools and techniques, taking advantage of the automation provided by machine learning approaches, and user knowledge through interactions with visual representations to help people understand and take advantage of complex and massive data collections. In the past years, his primary focus has been on designing and developing visual analytics techniques for the general public to advance the concept of data democratization, promoting unconstrained access to data analysis and widening the analytic capability of lay users in transforming data into insights. He is an IEEE member.
\end{IEEEbiography}




\end{document}